\documentclass[journal]{IEEEtran}

\usepackage{makecell} 
\usepackage{fontspec}
\usepackage{varwidth} 
\usepackage{lipsum}
\usepackage{multicol}
\usepackage{multirow}
\usepackage{booktabs}
\usepackage{mwe}
\usepackage{enumerate}
\newcommand{\subparagraph}{}
\usepackage{titlesec}
\usepackage{colortbl}
\usepackage{verbatim}
\usepackage{overpic}
\usepackage{indentfirst}
\usepackage[dvipsnames,HTML]{xcolor}
\usepackage{adjustbox}

\usepackage{caption}
\usepackage{subcaption}
\usepackage{booktabs}
\usepackage{tabularx}                   
\usepackage{threeparttable}
\usepackage{multicol}
\usepackage{multirow}
\usepackage{amssymb}
\usepackage{amsfonts}
\usepackage{graphicx}
\usepackage{float}
\usepackage{soul}
\usepackage{comment}
\usepackage{amsmath}
\usepackage{threeparttable}

\begin{document}
%
\title{Motion comfort and driver feel: An explorative study about their relation in remote driving}

\author{Georgios~Papaioannou*,
        Lin~Zhao, Mikael Nybacka, Jenny Jerrelind, Riender Happee and Lars Drugge
\thanks{Georgios Papaioannou and Riender Happee are with the Cognitive Robotics Department, Delft University of Technology (E-mail:g.papaioannou@tudelft.nl).
Lin Zhao, Mikael Nybacka, Jenny Jerrelind and Lars Drugge are with the Department of Engineering Mechanics, KTH Royal Institute of Technology, Teknikringen 8, SE-100 44, Stockholm, Sweden.}
\thanks{Manuscript received April 19, 2005; revised August 26, 2015.}}

\markboth{ARXIV PREPRINT, SUBMITTED TO IEEE T-ITS IN APR. 2023}%
{Shell \MakeLowercase{\textit{et al.}}: Bare Demo of IEEEtran.cls for IEEE Journals}

\maketitle

\begin{abstract}

Teleoperation is considered as a viable option to control fully automated vehicles (AVs) of Level 4 and 5 in special conditions. 
However, by bringing the remote drivers in the loop, their driving experience should be realistic to secure safe and comfortable remote control. 
Therefore, the remote control tower should be designed such that remote drivers receive high quality cues regarding the vehicle state and the driving environment. 
In this direction, the steering feedback could be manipulated to provide feedback to the remote drivers regarding how the vehicle reacts to their commands. 
However, until now, it is unclear how the remote drivers' steering feel could impact occupant's motion comfort. 
This paper focuses on exploring how the driver feel in remote (RD) and normal driving (ND) are related with motion comfort. 
More specifically, different types of steering feedback controllers are applied in (a) the steering system of a Research Concept Vehicle-model E (RCV-E) and (b) the steering system of a remote control tower. An experiment was performed to assess driver feel when the RCV-E is normally and remotely driven. 
Subjective assessment and objective metrics are employed to assess drivers' feel and occupants' motion comfort in both remote and normal driving scenarios. 
The results illustrate that motion sickness and ride comfort are only affected by the steering velocity in remote driving, while throttle input variations affect them in normal driving.
The results demonstrate that motion sickness and steering velocity increase both around 25$\%$ from normal to remote driving. 

\end{abstract}


\section{Introduction}

Automated vehicles (AVs) acceptance and employment is deterred by major concerns related to motion comfort \cite{AtifahSaruchi2022} (the term referring to both motion sickness and ride comfort), and their ability to be controlled in special conditions \cite{Singh2021}. 

AVs will be able to handle most of the manoeuvres in urban environments. 
However, a multitude of factors, (i.e. bad weather \cite{Vargas2021}, low sensor perception \cite{VanBrummelen2018}, difficult scenarios \cite{Shetty2021}, constructions, public events, traffic accidents, etc.) can lead to limited sensor recognition and a mismatch of the actual road with outdated high precision maps. 
Hence, difficulties might rise for AVs to handle such conditions, leading to stranded vehicles and accidents, which will eventually risk AVs deployment. 
Meanwhile, the ability to engage in other activities during the ride is considered by consumers as one of the key reasons for AVs adoption \cite{Mosquet2015}. 
However, the engagement in non-driving activities will provoke occupants' motion sickness (MS), deteriorating their overall motion comfort and thereby risking AVs acceptance.
Therefore, research has been recently conducted towards both challenges. 
However, we found no work investigating their interaction. 

The main countermeasure overcoming AVs difficulty to handle special conditions is the remote driving (RD) technology or teleoperation \cite{Zhang2020}.  
RD technology could be an effective way to provide a backup system to AVs, smoothening the transition phase towards the employment of fully automated vehicles. 
With the introduction of 4G and 5G communication technology, stable RD can be realized by using high-resolution and low latency video.
However, the remote drivers' driving behaviour will be affected by a lack of physical (e.g. vestibular) motion cues regarding the vehicle behaviour.
As a result, the accelerating, braking and steering behaviour might become more aggressive and with more sudden jerks, hampering occupants' motion comfort.
Therefore, it is crucial to identify the relation between remote drivers' driving feel and occupant's motion comfort to be able to improve both.

Conventional road vehicles are designed to be driven by and interact with the human driver/operator. 
In the vehicle-driver closed-loop system, the driver receives important feedback from the vehicle motion, including the vehicle’s velocity and acceleration, posture, and steering feel.
Traditionally,  drivers get most of these through haptic and kinesthetic feedback via the steering system, the seat, and the brake/gas pedal. 
Among these, the vehicle steering system includes key information (e.g. the vehicle speed, the tyre forces, and the road condition \cite{Nguyen2009}), which the remote driver needs to properly control the vehicle.
Hence, the modification of the remote steering system could be a direct way to transmit real world information to the remote driver. 
Nevertheless, on contrary with the traditional vehicle steering system, the remote control tower steering system might not be equipped with the steering column, rack, and other components. 
Therefore, it is necessary to provide artificial cues to the remote drivers to enhance their steering behaviour \cite{Jensen201160}.
However, it is unclear how these changes could affect occupants' motion comfort. 

The remote steering system can be compared to a Steer-by-Wire (SBW) steering system. 
SBW are being carefully designed to create a desired steering feel, transmitting the aligning torque resulting from tyre-road interaction.
Hence, the different kinds of steering feedback models that have been developed for SBW systems \cite{Mortazavizadeh20206,Chugh202058a,Balachandran201520,Fankem201452} can be employed also for remote steering systems. 
The difference with the normal driven vehicles is that in RD there is limited to no steering feedback information about how the vehicle reacts to the drivers' commands.
Additionally, the visual and motion cues are significantly different, where vision generally sees a reduced resolution and field of view, and fixed based setups do not elicit any mechanical motion cues. 
Thus, the requirements of the steering feedback information in RD could be principally different from normal driving.
Till now, the authors' previous study \cite{Lin2022a}, presented a novel experiment to test the drivers' feel in normal and remote driving. 
Through this experiment, they outlined that the requirements of the amplitude of steering feedback force and returnability in remote driving are lower than that of normal driving, while they identified that it was more difficult and less safe to operate the vehicle remotely than normally.
However, despite these critical conclusions, there was no investigation about how drivers' feel could affect occupants' motion comfort in normal and remote driving.



\begin{figure}[!h]
\centering
\begin{subfigure}[b]{0.8\linewidth}
	\centering 
	\includegraphics[width=\linewidth]{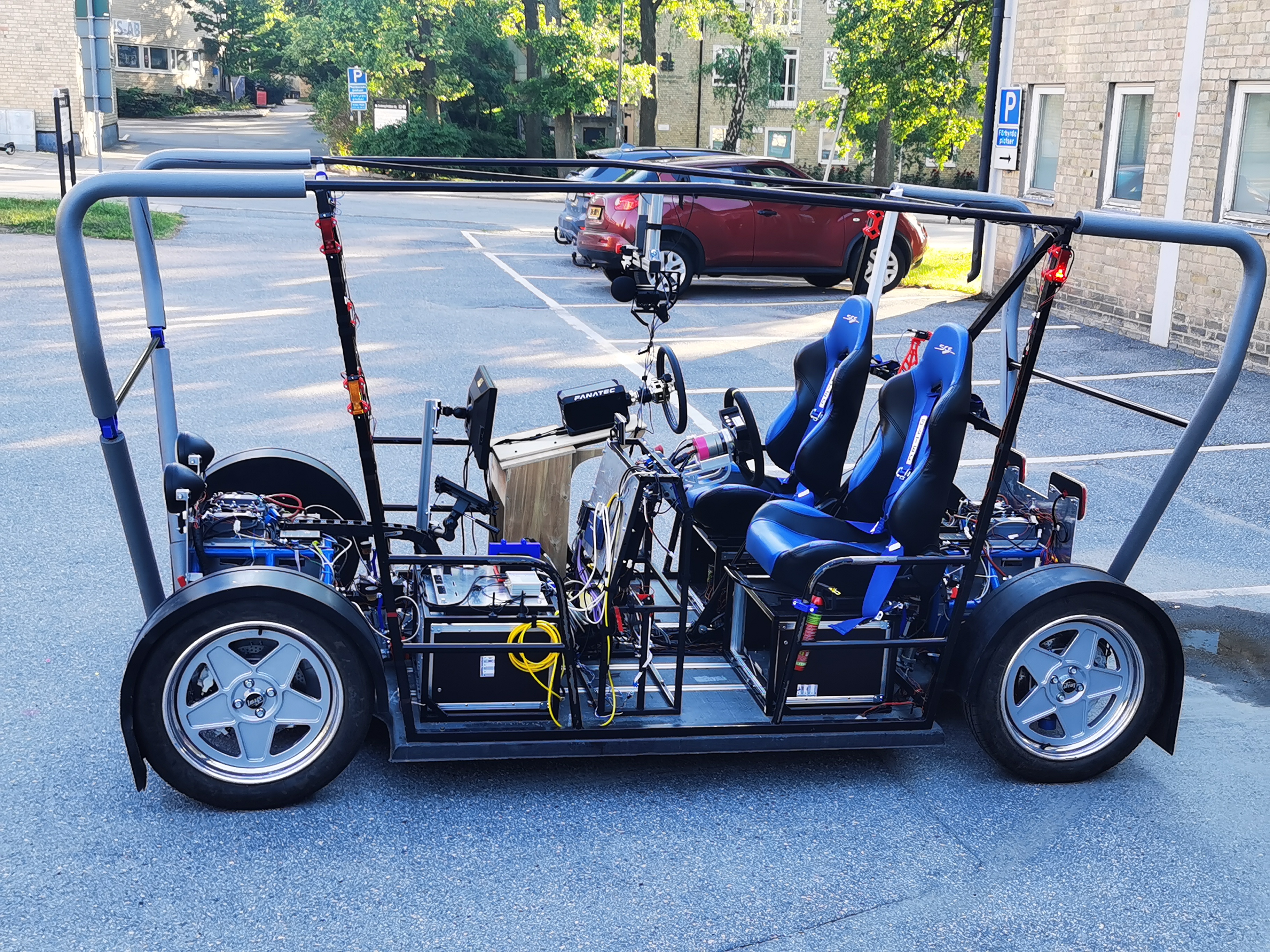}
	\caption{\label{fig:RCVE}}
\end{subfigure}
\hfil
\begin{subfigure}[b]{0.8\linewidth}	
	\centering 
	\includegraphics[width=\linewidth]{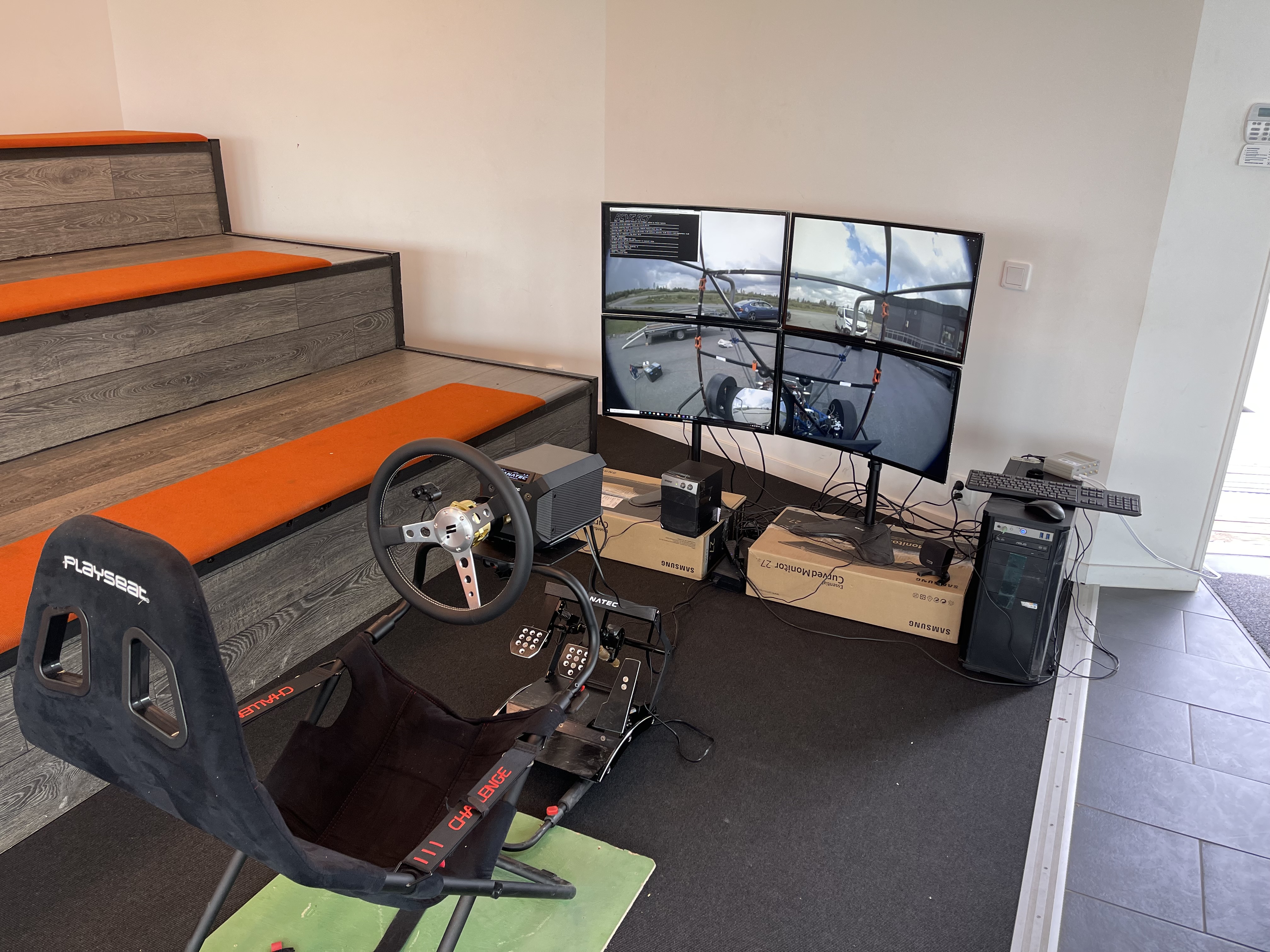}
	\caption{\label{fig:control_tower}}	
\end{subfigure}
\caption{(a) Research Concept Vehicle-model E (RCV-E).  (b)  Remote Control Tower (RCT) \label{fig:ModelComparison}}
\end{figure}

In this direction, this work exploits the data collected from the novel experiment with human drivers as partially reported in authors' previous work \cite{Lin2022a}, and explores in depth the relation of motion comfort with drivers' feel. 
More specifically, we used the Research Concept Vehicle-model E (RCV-E, Figure \ref{fig:RCVE}) with a remote control tower (RCT, Figure \ref{fig:control_tower}), where they are equipped with the same steering feedback interface (Fanatec kits) in RCV-E and RCT, and it can provide drivers with high fidelity steering feedback force. The human drivers tested the steering feel both during normal and remote driving. 
To study the effect of different steering conditions, three different steering feedback models are applied to provide different feedback and affect the drivers' steering feel during normal (ND) and remote (RD) driving. 
Subjective assessment metrics are extracted through questionnaires during a specific manoeuvre, while objective metrics are also calculated to acquire comprehensive data regarding drivers' feel. 
Using the data obtained from sensors placed on the RCV-E, occupants' motion sickness (MS) and ride comfort (RC) are assessed with objective metrics. 
The results aim to explore the relation of motion comfort with remote drivers' feel and pave the path for properly designing steering feedback in remote driving with the consideration of motion comfort.

The novel contributions of our paper jointly outline the criticality of considering motion comfort in the design of remote control systems:
\begin{itemize}
    \item We identify differences between normal and remote drivers' overall steering feel and its components through the subjective assessment metrics, and we suggest guidelines regarding the improvement of the subjective assessment questionnaire. 
    \item We prove for the first time the hypothesis that motion sickness is increased from normal to remote driving, while steering velocity and travelled distance are also increased. 
    \item We validate the relation of ride comfort and motion sickness with steering velocity and throttle input in normal driving, while we demonstrate that steering velocity is the dominant factor in remote driving.
    \item We substantiate the correlation of ride comfort and motion sickness with different subjective driver feel assessment metrics, and we capture for the first time the contradictory impact that the subjective driver feel has on ride comfort and motion sickness in remote and normal driving. 
\end{itemize}

To that end, this paper is structured as follows: first, the
methods used to assess the steering feel and the motion comfort are presented; secondly, the experimental scenarios are described together with the experimental setup; then, the results are presented and discussed; finally, conclusions are extracted.

\section{Steering controllers} 

This work focuses on exploring how steering feel in RD affects occupants' motion comfort.
For this, three steering feedback controllers are applied to the steering systems in the RCV-E and the remote control tower: the modular model (MF), the physical model (PF) and the no-feedback (NF).
More details regarding these models can be found in the literature \cite{Lin2022a}, where they are extensively studied regarding their impact on remote drivers' steering feel.
The emphasis of the current paper is to unravel the relation of between remote drivers' feel and occupants' motion comfort, rather than exploring the impact of different feedback controllers on motion comfort. 
To that end, the different feedback controller scenarios are considered as cases that provoke different drivers' steering feel, and are employed to investigate the overall relation of the drivers' feel with motion comfort. 

\section{Driver feel and motion comfort assessment}

\subsection{Subjective assessment of driver feel}

To subjectively assess the driver feel, a three-level based questionnaire is designed, as shown in Table \ref{tab:SA_Metrics} \cite{Lin2022a,Mandhata2012}.  
The first level (SA01) is about the overall assessment, while the second level questions concern the driver's perceived safety (SA10), steering wheel characteristic feel (SA20) as well as confidence and control (SA30).
The third level delves into these three points with two additional questions for SA10, SA20 and SA30 respectively. 
Each question has a grading scale  from 0 to 5 with a step of 0.25.  

\begin{table*}[ht]
\caption {Subjective driver feel assessment questionnaire.}  
\label{tab:SA_Metrics} 
\begin{center}
\begin{tabular}{ | c | c| c|}
    \hline
    \thead{Level 1} & \thead{Level 2} & \thead{Level 3} \\ 
    \hline
    \multirow{6}{*}{\makecell[c]{ \\ \\ \\ \\ \\ \\ \\ SA01. Overall assessment\\\scriptsize  0\ldots 1\ldots 2\ldots 3\ldots 4 \ldots 5 \\ \scriptsize  \emph{Bad} (0) $\rightarrow$ Neutral $\rightarrow$  \emph{Good} (5)}} & 
      
    \multirow{2}{*}{\makecell[c]{\\ \\ \footnotesize SA10. Safety assessment
    \\ \scriptsize 0\ldots 1\ldots 2\ldots 3\ldots 4 \ldots 5 
    \\ \scriptsize  \emph{Unsafe} (0) $\rightarrow$ Neutral $\rightarrow$\emph { Safe} (5)}}  
      
    & \makecell[c]{\footnotesize SA11. Steering feedback support to\\ \footnotesize control the vehicle through manoeuvre:\\ \scriptsize  0\ldots 1\ldots 2\ldots 3\ldots 4 \ldots 5 \\  \scriptsize  \emph{Little support} (0) $\rightarrow$ Neutral $\rightarrow$ \emph{High support} (5)}\\ \cline{3-1}

    &  & \makecell[c]{\footnotesize SA12 Steering feedback\\ \footnotesize communication of the vehicle behaviour:\\ \scriptsize  0\ldots 1\ldots 2\ldots 3\ldots 4 \ldots 5 \\  \scriptsize  \emph{Bad} (0) $\rightarrow$ Neutral $\rightarrow$  \emph {Good} (5)}\\\cline{2-3}

    & \multirow{2}{*}{\makecell[c]{\\ \footnotesize  SA20. Steering wheel  characteristic  feel 
    \\\scriptsize  0\ldots 1\ldots 2\ldots 3\ldots 4 \ldots 5 
    \\  \scriptsize  \emph{Unrealistic} (0) $\rightarrow$ Neutral $\rightarrow$ \emph{Realistic} (5)}}  
      
    & \makecell{\footnotesize SA21. Level of feedback force:\\
      \scriptsize  0\ldots 1\ldots 2\ldots 3\ldots 4 \ldots 5 \\  \scriptsize  \emph{Too small} (0) $\rightarrow$ Neutral $\rightarrow$ \emph {Too large} (5)}\\ \cline{3-1}
      
    &  & \makecell{\footnotesize SA22. Returnability of steering wheel to centre:\\ \scriptsize  0\ldots 1\ldots 2\ldots 3\ldots 4 \ldots 5  \\   \scriptsize  \emph{Too slow} (0) $\rightarrow$ Neutral $\rightarrow$ \emph {Too quick} (5)}\\\cline{2-3} 
      
    &  \multirow{2}{*}{\makecell{\\ \\  \footnotesize SA30. Confidence and control
    \\ \scriptsize  0\ldots 1\ldots 2\ldots 3\ldots 4 \ldots 5  
    \\ \scriptsize  \emph{Unconfident} (0) $\rightarrow$ Neutral $\rightarrow$ \emph{Confident} (5)}}
 
    & \makecell{\footnotesize SA31. Your assessment of the degree of \\\footnotesize success of accomplishing the task is:\\ \scriptsize  0\ldots 1\ldots 2\ldots 3\ldots 4 \ldots 5  \\  \scriptsize  \emph{Failure} (0) $\rightarrow$ Neutral $\rightarrow$ \emph{Success} (5)}\\ \cline{3-1}
      
    &    & \makecell{\footnotesize SA32. Your assessment of the difficulty \\ \footnotesize of accomplishing the task is:\\ \scriptsize  0\ldots 1\ldots 2\ldots 3\ldots 4 \ldots 5  \\  \scriptsize  \emph{Easy} (0) $\rightarrow$ Neutral $\rightarrow$ \emph{Difficult} (5)}\\\cline{2-3} 
      
      
      
        
  \hline
\end{tabular}
\end{center}
\end{table*}

\subsection{Objective assessment of motion comfort}

ISO-2631:1998 \cite{ISO2631} provides objective guidelines for measurement and evaluation of human exposure to whole-body mechanical vibration and repeated shock. Here we use and extend these guidelines to derive two comfort metrics being:
\begin{enumerate}
    \item Ride Comfort (RC) emphasizing the higher frequencies (mainly above 1 Hz).
    \item Motion Sickness (MS) emphasizing the lower frequencies (mainly below 1 Hz).
\end{enumerate}
Both metrics apply frequency weighting to 6 degrees of freedom motion including three dimensional translation and  three dimensional rotation of the seat or the head. The Ride Comfort (RC) is expected to capture general motion (dis)comfort due to vibration and abrupt motion and is deemed relevant to active motion (driving) and passive motion (being driven). The second measure is suitable for passive motion (being driven).

According to the standard, comfort is assessed by combining the root mean square (RMS) values of weighted accelerations ($RC_{W_{i}}$), translational and rotational, measured at the vehicle's centre of gravity.
More specifically, the RMS value of each acceleration is calculated as follows: 

\begin{equation}
\label{eq:accel}
	RC_{W_{i}}=  \bigg( \frac{1}{t}  \int_{0}^{t} a_{i_W}^2 d\tau \bigg)^{\frac{1}{2}}
\end{equation}

\noindent where $i$ is the acceleration type, either translational ($\ddot{x}$, $\ddot{y}$ and $\ddot{z}$) or rotational ($i$= $rx$ for $\ddot{\phi}$, $ry$ for $\ddot{\theta}$ and $rz$ for $\ddot{r}$) as defined in the standart \cite{ISO2631}, while $a_{W_i}$ stands for the weighted accelerations in the time domain.
After multiplying each of the $RC_{iW_{rms}}$ by appropriate factors ($k_i$), they are all summed and the overall comfort metric is calculated: 

\begin{equation}
\label{eq:RC}
RC = \bigg( \sum_{i=1}^{6} k_i^2 RC_{W_{i}}^2 \bigg)^{1/2}
\end{equation}

\noindent where $k_i$ is the multiplying factor for each term ($i$=$x$, $y$, $z$, $rx$, $ry$ and $rz$) which can be found in ISO-2631 \cite{ISO2631}.
As far as the weighting of the accelerations is concerned, they are calculated as follows:

\begin{equation}
\label{eq:weighting}
    A_{w_{i}} =  WP_{i_{1}} * WA_{i_{1}} * A_{i}
\end{equation}

\noindent where \textbf{$A_i$} are the frequency domain accelerations ($a_i(t)$); \textbf{$A_{w_{i}}$} are the frequency domain weighted accelerations; \textbf{$WP$} and \textbf{$WA$} are the principal and additional frequency weightings used. 

\begin{figure}[ht!]
\centering
\begin{subfigure}{0.8\linewidth}
  \centering 
  \includegraphics[width=\linewidth]{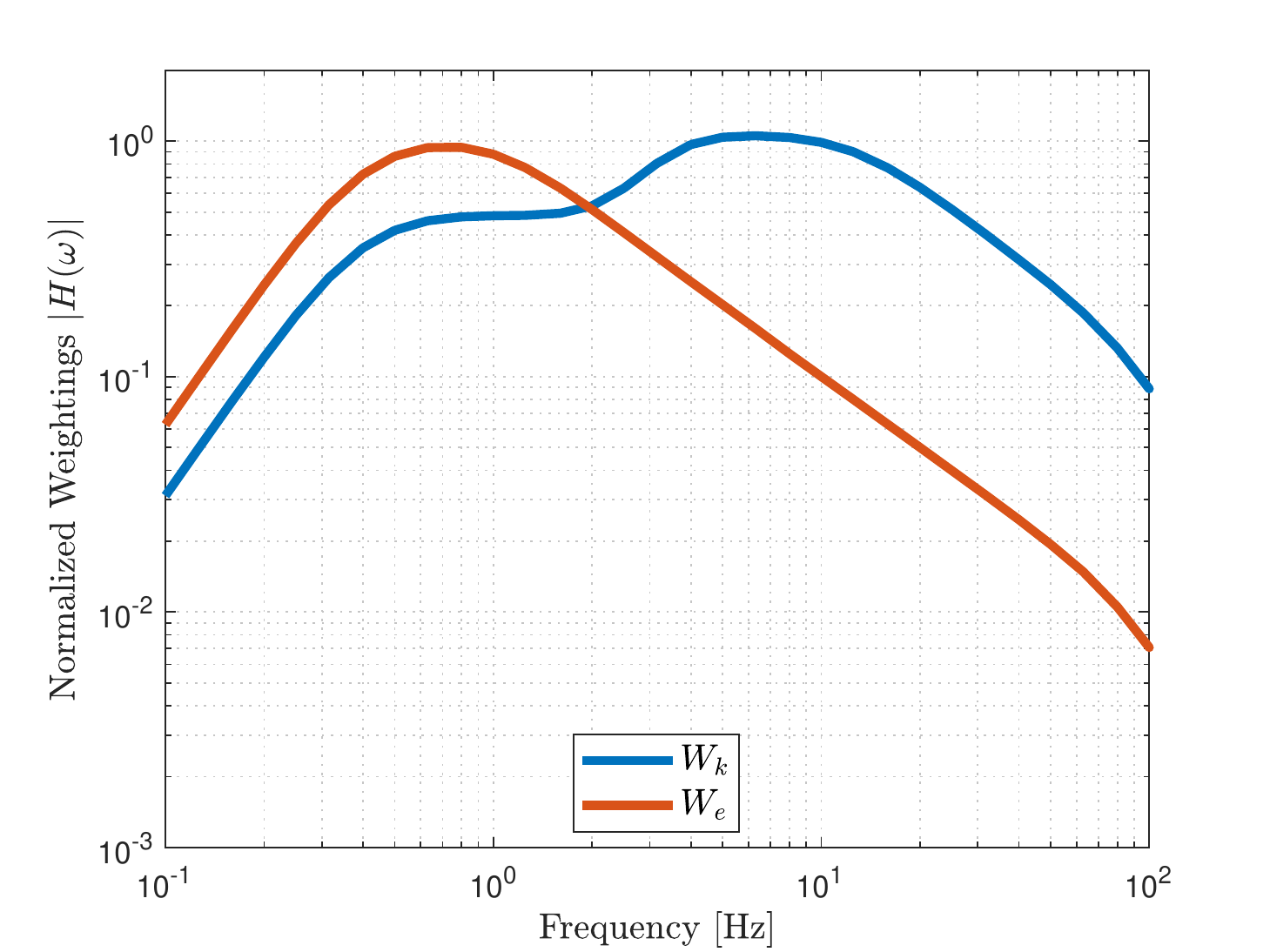}
  \label{fig:Filters_WC}
\end{subfigure}
\begin{subfigure}{0.8\linewidth}
  \centering 
  \includegraphics[width=\linewidth]{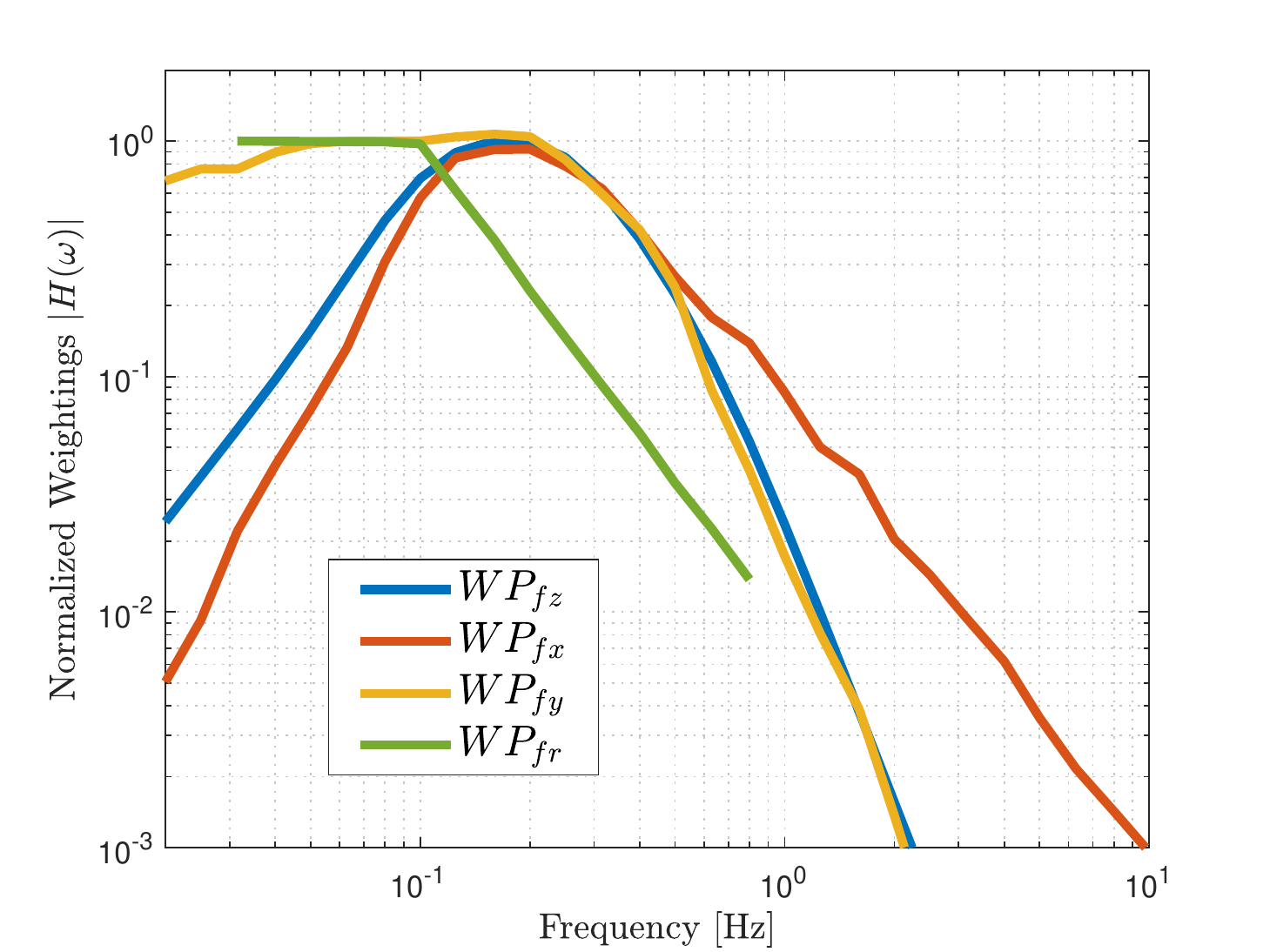}
  \label{fig:Filters_WF}
\end{subfigure}
\caption{Weighting filters regarding ride comfort and motion sickness. \label{fig:Filters}}
\end{figure}

Equation \ref{eq:RC} is used with two different sets of weighting filters for the translational and rotational accelerations to objectively assess RC and MS.
The combinations of filters used for each case (RC and MS metrics) are illustrated in Table \ref{tab:filters}, while all filters used in this work are displayed in Figure \ref{fig:Filters}.
Regarding RC, according to ISO-2631, the principal filter for the z-direction ($WP_k$) and the additional filter ($WA_e$) for all the rotational vibrations are used. 
No filter is used in x and y direction according to the standard \cite{ISO2631}. 
As far as MS is concerned, ISO-2631 lacks appropriate weighting filters for the horizontal and rotational vibrations despite their importance in MS accumulation.
Therefore, more filters from the literature are employed. 
More specifically, $WP_{f_x}$, $WP_{f_y}$ and $WP_{f}$ are used for the x, y and z direction, while $WA_{f_r}$ is used for all the rotational vibrations.
The longitudinal acceleration weighting filter ($WP_{f_x}$) is approximately designed according to Griffin et al. \cite{Griffin2002}, the lateral acceleration weighting filter ($WP_{f_y}$) is extracted from Donohew et al. \cite{Donohew2004} and the rotational vibration weighting filter ($WA_{f_r}$) is designed based on Howarth et al. \cite{Howarth2003}.

\begin{table}[h!]
    \centering
    \caption{Applied weighting filters in translational and rotational accelerations for the assessment of ride comfort and motion sickness.}
    \label{tab:filters}    
    \begin{tabular}{|c|c|c|c|c|c|c|}
         \hline
         \multicolumn{7}{|c|}{\multirow{2}{*}{\textbf{Weighting Filters}}} \\
         \multicolumn{7}{|c|}{} \\         
         \hline
         \multirow{2}{*}{\textbf{Application}}  & \multicolumn{6}{c|}{\textbf{Vibration}}\\
         \cline{2-7} 
                                 &  \textbf{x}  & \textbf{y}  & \textbf{z}  &  \textbf{rx}  & \textbf{ry}  & \textbf{rz} \\     
         \hline
         Ride comfort (RC)       & - & - & $WP_k$              & \multicolumn{3}{c|}{$WA_e$} \\ 
         \hline
         Motion sickness (MS)    &  $WP_{f_x}$   & $WP_{f_y}$   & $WP_{f_z}$   & \multicolumn{3}{c|}{$WA_{f_r}$}  \\ 
         \hline
    \end{tabular}

\end{table}


\subsection{Multidimensional human body model}

In this work, the vehicle acceleration measurements ($\ddot{x}$, $\ddot{y}$, $\ddot{z}$, $\ddot{r}$, $\ddot{\phi}$ and $\ddot{\theta}$) from the Xsens IMU are transmitted to the occupants' head ($\ddot{x}_h$, $\ddot{y}_h$, $\ddot{z}_h$, $\ddot{r}_{h}$, $\ddot{\phi}_{h}$ and $\ddot{\theta}_{h}$) using a multidimensional human body model \cite{Papaioannou2021b}. 
This takes into account body induced oscillations, and head rotational responses to seat translational and rotational accelerations.

\section{Experiment Design}

\subsection{Experiment Setup}

\begin{figure}[H]
   \centering
    \includegraphics[width=1.0\linewidth]{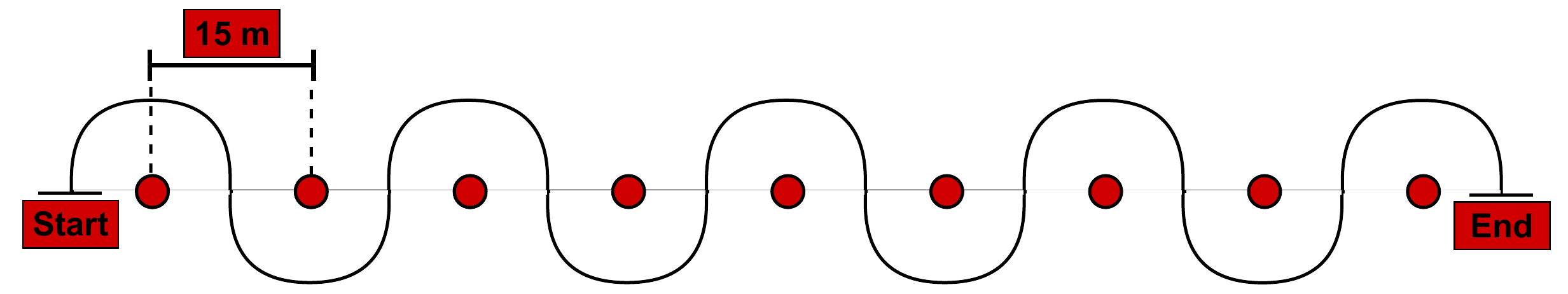}
    \caption{Test arrangement of the slalom.}
    \label{fig:slalom}
\end{figure}

A slalom manoeuvre, as shown in Figure \ref{fig:slalom}, is designed for the experiment.
More specifically, nine cones were placed with distance of 15 m between each. 
Considering safety aspects, the maximum speed in the experiment was limited to 18 km/h.
For the data acquisition, the Fanatec Kit was used to measure the steering angle and throttle/brake pedal movement data, and it was also used to generate the steering feedback force. 

\subsection{Experiment Protocol}

In total, five participants were involved and all of them have driving licenses and driving experience in ND. 
Three of the participants did not have any RD experience, but the last two are regularly engaged in RD at Einride, a company for automated heavy vehicles. 
Each participant assessed the driving feel for each steering feedback model (PF, MF and NF - three in total), in both ND and RD. 
The ND scenario was firstly conducted to let drivers be familiar with the RCV-E, and then the RD scenario followed. 
In total, 15 cases were studied (i.e., 5 different drivers testing 3 different steering feedback models) for each scenario, i.e. both ND and RD.

Before the initiation of the experiment, the drivers were asked to close their eyes for two minutes to stabilise the brain wave activity. 
Then, to become familiar with the manoeuvre, the drivers practiced the manoeuvre for four laps using each of the steering feedback models. 
Afterwards, the formal test followed by testing each feedback model. 
In order to alleviate the influence of the model's testing sequence on the drivers' assessment, the sequence varied among the drivers.
This was also conducted during the initial training. 
After the formal experiment for each steering model, the drivers were asked to answer immediately the questions for each model. 
RD also had the same procedure.  
 


\section{Results \& Discussion}

In total, 15 cases are studied (i.e., 5 different drivers testing 3 different steering feedback models) for each scenario, i.e. normal (ND) and remote (RD) driving. 
All drivers completed the different cases successfully without loss of control or collisions.
The data obtained are analysed in the following way: 
\begin{itemize}
    \item The correlation between the subjective metrics (Table \ref{tab:SA_Metrics}) is investigated to understand the differences in driver feel between ND and RD, and outline any redundant metrics that can be excluded from future analysis (Table \ref{tab:subj_correlation}). 
    \item Objective metrics are used to capture the impact of the driving behaviour and steering feel on motion comfort during RD and ND. 
    More specifically, ride comfort (RC), motion sickness (MS), steering velocity (SV), throttle input velocity (TI) and distance travelled (DT) are compared between ND and RD through boxplots (Figures \ref{fig:RC_MS_box}-\ref{fig:DT_SV_DS_box}).
    SV and TI are studied using the RMS value of the corresponding measurements over one drive consisting of multiple slaloms (Figure \ref{fig:slalom}), while the total distance travelled (DT) is calculated based on the IMU data. 
    The DT metric is considered since when drivers adopt a wider slalom longer distance will be covered affecting MS and RC.    
    Ride comfort and motion sickness are represented by RC and MS as defined in the previous sections. 
    \item The correlation of RC and MS with SV and TI is investigated in both ND and RD (Figures \ref{fig:MS_RC_DT_SV_S}-\ref{fig:MS_RC_DT_SV}) to unravel their relationship.
    The analysis is conducted for two time periods: (a) the first 0-2 s where the vehicle accelerates, and (b) the complete slalom.
    For the latter, a multiple regression model is used to validate the correlation results (Table \ref{tab:multiple_regression1}).
    \item RC and MS are investigated in terms of their correlation (Table \ref{tab:RSquareOfCorrelation}) with different steering feel subjective assessment metrics (Table \ref{tab:SA_Metrics}). 
    Cases highly correlated with MS or RC are investigated in detail by presenting the fitted curves (Figures \ref{fig:RC_MS_SA01_S} and \ref{fig:RC_MS_SA31_S}). 
\end{itemize}

\subsection{Correlation between subjective steering feel metrics}

Before investigating in depth the correlation of the subjective steering feel metrics (SA) with MS and RC, the dependency of the SA metrics is explored.
For this, a method to yield the correlation of the objectives and unravel potential redundancies is employed \cite{Lindroth2010}.  
The correlation coefficient of two objective functions ($SA_i$ and $SA_j$, Table \ref{tab:SA_Metrics}) over a set of decision vectors $Y$ = ${x^1,...,x^N}$, i.e. the $N$ driving scenarios in ND or RD, is defined as: 

\begin{equation}
    \rho _{SA_i,SA_j} = \frac{S_{i,j}}{S_{i,i}S_{j,j}} \in [-1,1]
    \label{eq:rho_cor}
\end{equation}

\noindent where $i,j$ = $[01,~10,~11,~12,~20,~21,~22,~30,~31,~32]$ and $S_{i,j}$ is defined:


\renewcommand\tabcolsep{12.0pt}
\begin {table*}[h!]
\centering
\caption {Correlation coefficient  ($\rho_{SA_i,SA_j}$) between the steering feel subjective assessment metrics (SA) in ND and RD. The colors depict the following levels of $\rho$: (a) red: $\rho$ = 1 (Utopia), (b) orange: 1 $> \rho \ge$ 0.94 (high), (c) green: 0.94 $> \rho >$ 0.80 (very good), (d) yellow: 0.80 $\ge \rho \ge$ 0.50 (good), and (e) blue: $\rho >$ 0.50 (bad). \label{tab:subj_correlation}}
\begin{tabular}{ccccccccccc}
\hline
\multicolumn{11}{c}{\textbf{\multirow{2}{*}{Normal Driving}}} \\
 \\
 \hline
 & \textbf{SA01} & \textbf{SA10} & \textbf{SA11} & \textbf{SA12} & \textbf{SA20} & \textbf{SA21} & \textbf{SA22} & \textbf{SA30} & \textbf{SA31}  & \textbf{SA32}\\ 
\hline
\textbf{SA01}         & \cellcolor[HTML]{C00000}1.00 & \cellcolor[HTML]{ED7D31}0.96 & \cellcolor[HTML]{70AD47}0.85 & \cellcolor[HTML]{ED7D31}0.94 & \cellcolor[HTML]{ED7D31}0.97 & \cellcolor[HTML]{ED7D31}0.96 & \cellcolor[HTML]{FFC000}0.50 & \cellcolor[HTML]{70AD47}0.84 & \cellcolor[HTML]{FFC000}0.55 & \cellcolor[HTML]{5B9BD5}0.10 \\
\textbf{SA10}                & \cellcolor[HTML]{E7E6E6}0.00 & \cellcolor[HTML]{C00000}1.00 & \cellcolor[HTML]{ED7D31}0.94 & \cellcolor[HTML]{70AD47}0.84 & \cellcolor[HTML]{ED7D31}0.98 & \cellcolor[HTML]{ED7D31}0.94 & \cellcolor[HTML]{5B9BD5}0.47 & \cellcolor[HTML]{70AD47}0.85 & \cellcolor[HTML]{5B9BD5}0.44 & \cellcolor[HTML]{5B9BD5}0.19 \\
\textbf{SA11}                & \cellcolor[HTML]{E7E6E6}0.00 & \cellcolor[HTML]{E7E6E6}0.00 & \cellcolor[HTML]{C00000}1.00 & \cellcolor[HTML]{FFC000}0.70 & \cellcolor[HTML]{70AD47}0.90 & \cellcolor[HTML]{70AD47}0.88 & \cellcolor[HTML]{FFC000}0.54 & \cellcolor[HTML]{70AD47}0.85 & \cellcolor[HTML]{5B9BD5}0.18 & \cellcolor[HTML]{5B9BD5}0.43 \\
\textbf{SA12}                & \cellcolor[HTML]{E7E6E6}0.00 & \cellcolor[HTML]{E7E6E6}0.00 & \cellcolor[HTML]{E7E6E6}0.00 & \cellcolor[HTML]{C00000}1.00 & \cellcolor[HTML]{70AD47}0.90 & \cellcolor[HTML]{70AD47}0.88 & \cellcolor[HTML]{5B9BD5}0.43 & \cellcolor[HTML]{FFC000}0.72                         & \cellcolor[HTML]{FFC000}0.63 & \cellcolor[HTML]{5B9BD5}0.01 \\
\textbf{SA20}                & \cellcolor[HTML]{E7E6E6}0.00 & \cellcolor[HTML]{E7E6E6}0.00 & \cellcolor[HTML]{E7E6E6}0.00 & \cellcolor[HTML]{E7E6E6}0.00 & \cellcolor[HTML]{C00000}1.00 & \cellcolor[HTML]{ED7D31}0.96 & \cellcolor[HTML]{5B9BD5}0.47 & \cellcolor[HTML]{70AD47}0.84 & \cellcolor[HTML]{FFC000}0.56 & \cellcolor[HTML]{5B9BD5}0.09 \\
\textbf{SA21}                & \cellcolor[HTML]{E7E6E6}0.00 & \cellcolor[HTML]{E7E6E6}0.00 & \cellcolor[HTML]{E7E6E6}0.00 & \cellcolor[HTML]{E7E6E6}0.00 & \cellcolor[HTML]{E7E6E6}0.00 & \cellcolor[HTML]{C00000}1.00 & \cellcolor[HTML]{FFC000}0.56 & \cellcolor[HTML]{70AD47}0.86 & \cellcolor[HTML]{5B9BD5}0.45 & \cellcolor[HTML]{5B9BD5}0.20 \\
\textbf{SA22}                & \cellcolor[HTML]{E7E6E6}0.00 & \cellcolor[HTML]{E7E6E6}0.00 & \cellcolor[HTML]{E7E6E6}0.00 & \cellcolor[HTML]{E7E6E6}0.00 & \cellcolor[HTML]{E7E6E6}0.00 & \cellcolor[HTML]{E7E6E6}0.00 & \cellcolor[HTML]{C00000}1.00 & \cellcolor[HTML]{70AD47}0.82 & \cellcolor[HTML]{5B9BD5}0.17 & \cellcolor[HTML]{FFC000}0.53 \\
\textbf{SA30}                & \cellcolor[HTML]{E7E6E6}0.00 & \cellcolor[HTML]{E7E6E6}0.00 & \cellcolor[HTML]{E7E6E6}0.00 & \cellcolor[HTML]{E7E6E6}0.00 & \cellcolor[HTML]{E7E6E6}0.00 & \cellcolor[HTML]{E7E6E6}0.00 & \cellcolor[HTML]{E7E6E6}0.00 & \cellcolor[HTML]{C00000}1.00 & \cellcolor[HTML]{5B9BD5}0.19 & \cellcolor[HTML]{5B9BD5}0.33 \\
\textbf{SA31}                & \cellcolor[HTML]{E7E6E6}0.00 & \cellcolor[HTML]{E7E6E6}0.00 & \cellcolor[HTML]{E7E6E6}0.00 & \cellcolor[HTML]{E7E6E6}0.00 & \cellcolor[HTML]{E7E6E6}0.00 & \cellcolor[HTML]{E7E6E6}0.00 & \cellcolor[HTML]{E7E6E6}0.00 & \cellcolor[HTML]{E7E6E6}0.00 & \cellcolor[HTML]{C00000}1.00 & \cellcolor[HTML]{FFC000}0.67 \\
\textbf{SA32}                & \cellcolor[HTML]{E7E6E6}0.00 & \cellcolor[HTML]{E7E6E6}0.00 & \cellcolor[HTML]{E7E6E6}0.00 & \cellcolor[HTML]{E7E6E6}0.00 & \cellcolor[HTML]{E7E6E6}0.00 & \cellcolor[HTML]{E7E6E6}0.00 & \cellcolor[HTML]{E7E6E6}0.00 & \cellcolor[HTML]{E7E6E6}0.00 & \cellcolor[HTML]{E7E6E6}0.00 & \cellcolor[HTML]{C00000}1.00 \\
\hline
\multicolumn{11}{c}{\textbf{\multirow{2}{*}{Remote Driving}}} \\
 \\
\hline
 & \textbf{SA01} & \textbf{SA10} & \textbf{SA11} & \textbf{SA12} & \textbf{SA20} & \textbf{SA21} & \textbf{SA22} & \textbf{SA30} & \textbf{SA31}  & \textbf{SA32}\\ 
\hline
\textbf{SA01}         & \cellcolor[HTML]{C00000}1.00       & \cellcolor[HTML]{ED7D31}0.94 & \cellcolor[HTML]{FFC000}0.79 & \cellcolor[HTML]{5B9BD5}0.47 & \cellcolor[HTML]{5B9BD5}0.02 & \cellcolor[HTML]{5B9BD5}0.23 & \cellcolor[HTML]{5B9BD5}0.40 & \cellcolor[HTML]{FFC000}0.58 & \cellcolor[HTML]{FFC000}0.62 & \cellcolor[HTML]{5B9BD5}0.44 \\
\textbf{SA10}         & \cellcolor[HTML]{E7E6E6}0.00       & \cellcolor[HTML]{C00000}1.00 & \cellcolor[HTML]{70AD47}0.86 & \cellcolor[HTML]{FFC000}0.52 & \cellcolor[HTML]{5B9BD5}0.12 & \cellcolor[HTML]{5B9BD5}0.03 & \cellcolor[HTML]{5B9BD5}0.47 & \cellcolor[HTML]{FFC000}0.65 & \cellcolor[HTML]{FFC000}0.63 & \cellcolor[HTML]{FFC000}0.53 \\
\textbf{SA11}         & \cellcolor[HTML]{E7E6E6}0.00       & \cellcolor[HTML]{E7E6E6}0.00 & \cellcolor[HTML]{C00000}1.00 & \cellcolor[HTML]{5B9BD5}0.14 & \cellcolor[HTML]{5B9BD5}0.05 & \cellcolor[HTML]{5B9BD5}0.31 & \cellcolor[HTML]{5B9BD5}0.30 & \cellcolor[HTML]{FFC000}0.63 & \cellcolor[HTML]{5B9BD5}0.24 & \cellcolor[HTML]{FFC000}0.79 \\
\textbf{SA12}         & \cellcolor[HTML]{E7E6E6}0.00       & \cellcolor[HTML]{E7E6E6}0.00 & \cellcolor[HTML]{E7E6E6}0.00 & \cellcolor[HTML]{C00000}1.00 & \cellcolor[HTML]{5B9BD5}0.49 & \cellcolor[HTML]{5B9BD5}0.48 & \cellcolor[HTML]{5B9BD5}0.05 & \cellcolor[HTML]{5B9BD5}0.48 & \cellcolor[HTML]{70AD47}0.88 & \cellcolor[HTML]{5B9BD5}0.08 \\
\textbf{SA20}         & \cellcolor[HTML]{E7E6E6}0.00       & \cellcolor[HTML]{E7E6E6}0.00 & \cellcolor[HTML]{E7E6E6}0.00 & \cellcolor[HTML]{E7E6E6}0.00 & \cellcolor[HTML]{C00000}1.00 & \cellcolor[HTML]{5B9BD5}0.26 & \cellcolor[HTML]{5B9BD5}0.20 & \cellcolor[HTML]{5B9BD5}0.48 & \cellcolor[HTML]{5B9BD5}0.41 & \cellcolor[HTML]{5B9BD5}0.46 \\
\textbf{SA21}         & \cellcolor[HTML]{E7E6E6}0.00       & \cellcolor[HTML]{E7E6E6}0.00 & \cellcolor[HTML]{E7E6E6}0.00 & \cellcolor[HTML]{E7E6E6}0.00 & \cellcolor[HTML]{E7E6E6}0.00 & \cellcolor[HTML]{C00000}1.00 & \cellcolor[HTML]{5B9BD5}0.00 & \cellcolor[HTML]{5B9BD5}0.08 & \cellcolor[HTML]{FFC000}0.66 & \cellcolor[HTML]{5B9BD5}0.33 \\
\textbf{SA22}         & \cellcolor[HTML]{E7E6E6}0.00       & \cellcolor[HTML]{E7E6E6}0.00 & \cellcolor[HTML]{E7E6E6}0.00 & \cellcolor[HTML]{E7E6E6}0.00 & \cellcolor[HTML]{E7E6E6}0.00 & \cellcolor[HTML]{E7E6E6}0.00 & \cellcolor[HTML]{C00000}1.00 & \cellcolor[HTML]{5B9BD5}0.15 & \cellcolor[HTML]{5B9BD5}0.23 & \cellcolor[HTML]{5B9BD5}0.00 \\
\textbf{SA30}         & \cellcolor[HTML]{E7E6E6}0.00       & \cellcolor[HTML]{E7E6E6}0.00 & \cellcolor[HTML]{E7E6E6}0.00 & \cellcolor[HTML]{E7E6E6}0.00 & \cellcolor[HTML]{E7E6E6}0.00 & \cellcolor[HTML]{E7E6E6}0.00 & \cellcolor[HTML]{E7E6E6}0.00 & \cellcolor[HTML]{C00000}1.00 & \cellcolor[HTML]{5B9BD5}0.48 & \cellcolor[HTML]{FFC000}0.61 \\
\textbf{SA31}         & \cellcolor[HTML]{E7E6E6}0.00       & \cellcolor[HTML]{E7E6E6}0.00 & \cellcolor[HTML]{E7E6E6}0.00 & \cellcolor[HTML]{E7E6E6}0.00 & \cellcolor[HTML]{E7E6E6}0.00 & \cellcolor[HTML]{E7E6E6}0.00 & \cellcolor[HTML]{E7E6E6}0.00 & \cellcolor[HTML]{E7E6E6}0.00 & \cellcolor[HTML]{C00000}1.00 & \cellcolor[HTML]{5B9BD5}0.04 \\
\textbf{SA32}         & \cellcolor[HTML]{E7E6E6}0.00       & \cellcolor[HTML]{E7E6E6}0.00 & \cellcolor[HTML]{E7E6E6}0.00 & \cellcolor[HTML]{E7E6E6}0.00 & \cellcolor[HTML]{E7E6E6}0.00 & \cellcolor[HTML]{E7E6E6}0.00 & \cellcolor[HTML]{E7E6E6}0.00 & \cellcolor[HTML]{E7E6E6}0.00 & \cellcolor[HTML]{E7E6E6}0.00 & \cellcolor[HTML]{C00000}1.00\\
\hline
\end{tabular}
\end{table*}


\begin{multline}
    S_{i,j} = \frac{1}{N} \sum_{l\in N} \bigg( SA_i (x^l) - \frac{1}{N}\sum_{m \in N} SA_i(x^m)\bigg)* \\
    \bigg( SA_j (x^l) - \frac{1}{N}\sum_{m \in N} SA_j(x^m)\bigg)
    \label{eq:rho_S}
\end{multline}


The value of the pairwise correlation coefficients between two objective functions gives a measure of how similar the functions evaluate the set $Y$ = ${x^1,...,x^N}$. 
If the correlation is perfect, i.e., $|\rho_{SA_i,SA_j}|$ = 1,  one of the objectives or metrics is actually redundant and does not add any additional information.
According to Equations \ref{eq:rho_cor} and \ref{eq:rho_S}, the correlation coefficient was calculated regarding SA among the different driver responses both in ND and RD (Table \ref{tab:subj_correlation}). 
The different color cells indicates the level of correlation between the metrics, i.e. high, very good, etc., as described in the table.

Regarding ND (i.e., top part of Table \ref{tab:subj_correlation}), a few of the metrics from SA01-SA21 are significantly correlated ($\rho _{SA_i,SA_j}>$0.80) indicating potential redundancies. 
Firstly, as for the correlation of Level 1 and 2 metrics, SA01 (i.e., overall assessment - Level 1), is highly correlated with all the Level 2 metrics (i.e. SA10 - the safety assessment, SA20 - the steering wheel characteristic feel, and SA30 - confidence and control). 
In these cases,  $\rho _{SA_{01},SA_{10}}$ and $\rho _{SA_{01},SA_{20}}$ are $\sim$ 0.97, while $\rho _{SA_{01},SA_{30}}$ is $\sim$ 0.84.
This indicates that the overall drivers' steering feel assessment is predominantly related with their perception of safety and steering wheel characteristic feel, and secondarily with the perception of confidence and control. 
To that end, SA01 can be evaluated using multiple regression models based on SA10 and SA20, with which there is high correlation, or based on SA10, SA20 and SA30.

As far as the correlation of Level 2 and 3 metrics is concerned, SA10 is strongly correlated with SA11, i.e. the steering feedback support (Level 3), with $\rho _{SA_{10},SA_{11}}$ being $\sim$ 0.94. 
Based on this, the driver's safety assessment is mainly related with the driver perception about the steering feedback support levels (SA11) to control the vehicle rather than the steering feedback communication of vehicle behaviour (SA12).
At the same time, the drivers assessment regarding the steering wheel characteristic feel (SA20 - Level 2) is mainly related with the level of feedback force (SA21 - Level 3), but hardly correlated with the steering wheel returnability (SA22). 
Finally, the driver's confidence and control is highly correlated with the steering feedback support (SA11), the steering feedback communication (SA12), the level of feedback force (SA21) and the steering wheel returnability (SA22) rather than the assessed task success or difficulty (SA31 and SA32). 
Hence, SA31 and SA32 could be neglected since they do not add any additional information, while SA30 could be evaluated based on the above level 3 questions. 

Despite the significant correlations between metrics identified in ND, this is not the case in RD (i.e., bottom part of Table \ref{tab:subj_correlation}). 
On contrary with ND, the SA metrics illustrate very low correlation coefficients between them, except 1-2 cases. 
According to the correlation values, the overall driver feel assessment (SA01) is mainly and to some extent related with the drivers' perception of safety (SA10) and their confidence and control (SA30), respectively. 
Meanwhile, the importance of the steering wheel characteristic feel (SA20) was deteriorated compared to ND. 
Regarding the more in depth metrics of the questionnaire, the drivers' safety assessment and confidence are correlated with the steering feedback support (SA11) and the perceived task difficulty (SA32). 
At the same time, the remote drivers' confidence and control is related only with the perceived safety metrics (SA11 and SA12) on contrary with the normal driving scenario where the steering wheel characteristic feel is also critical. 
The above differences between the subjective driver feel assessment in ND and RD demonstrate the importance of further work to understand remote driver's steering feel in order for safe remote control systems to be designed. 



\subsection{RC and MS comparison between ND and RD}

\begin{figure}[h!]
\centering
\begin{subfigure}{0.75\linewidth}
  \centering 
  \includegraphics[width=\linewidth]{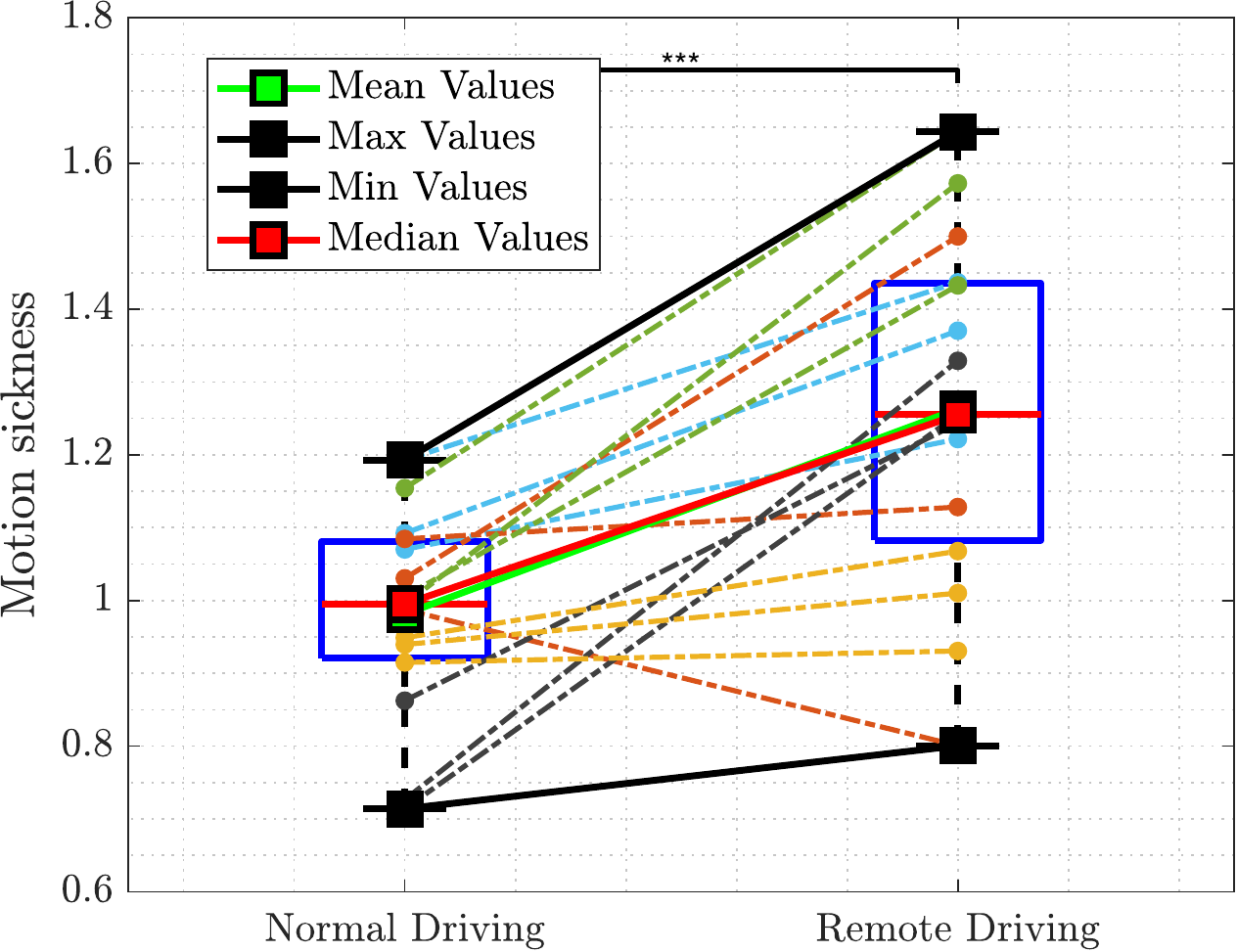}
  \caption{}
  \label{fig:MotionSicknessComparison}
\end{subfigure}

\begin{subfigure}{0.75\linewidth}
  \centering 
  \includegraphics[width=\linewidth]{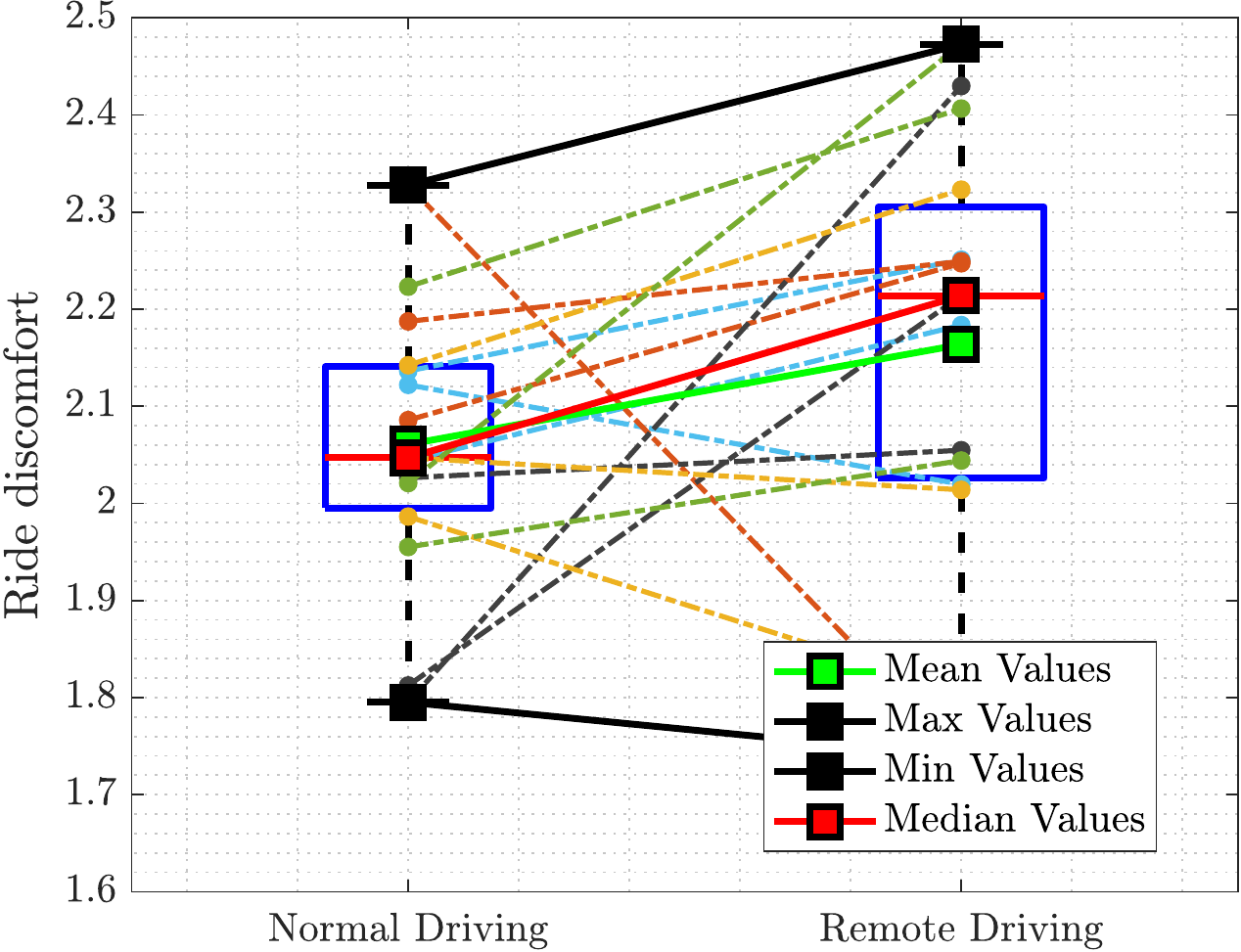}
  \caption{}
  \label{fig:RideComfortComparison}
\end{subfigure}
\caption{Comparison of (a) motion sickness and (b) ride discomfort in ND and RD. Paired sample t-test significance is presented by  $^{*}P\le0.05$,$^{**}P\le0.01$, and $^{***}P\le0.001$. [\textcolor[HTML]{77AC30}{\textbf{Driver 1}}, \textcolor[HTML]{4DBEEE}{\textbf{Driver 2}}, \textcolor[HTML]{EDB120}{\textbf{Driver 3}}, \textcolor[HTML]{D95319}{\textbf{Driver 4}}, \textcolor[HTML]{404040}{\textbf{Driver 5}}].}
\label{fig:RC_MS_box}
\end{figure}

\begin{figure}[h!]
\centering
\begin{subfigure}{0.75\linewidth}
  \centering 
  \includegraphics[width=\linewidth]{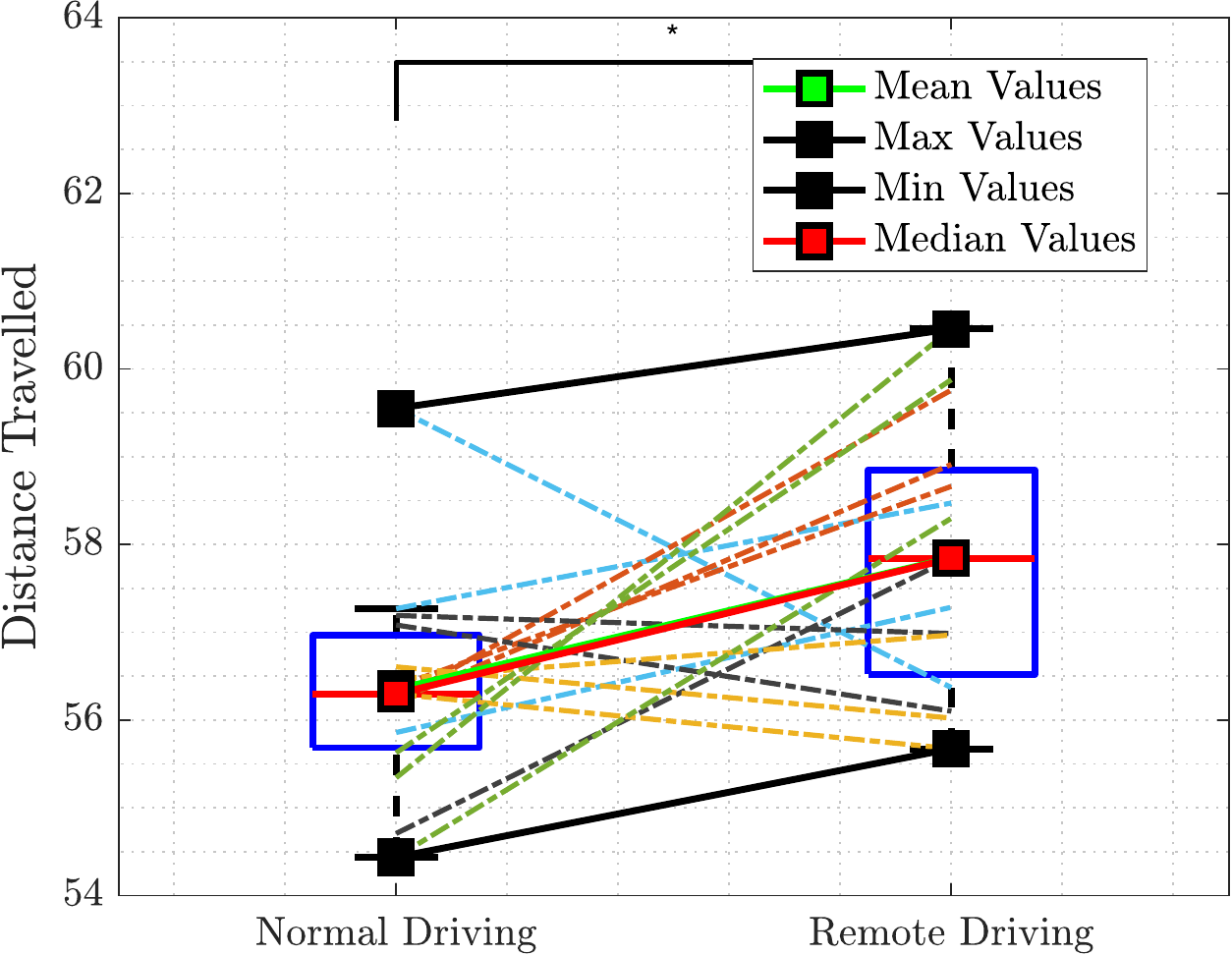}
  \caption{}
  \label{fig:DistanceComparison}
\end{subfigure}

\begin{subfigure}{0.75\linewidth}
  \centering 
  \includegraphics[width=\linewidth]{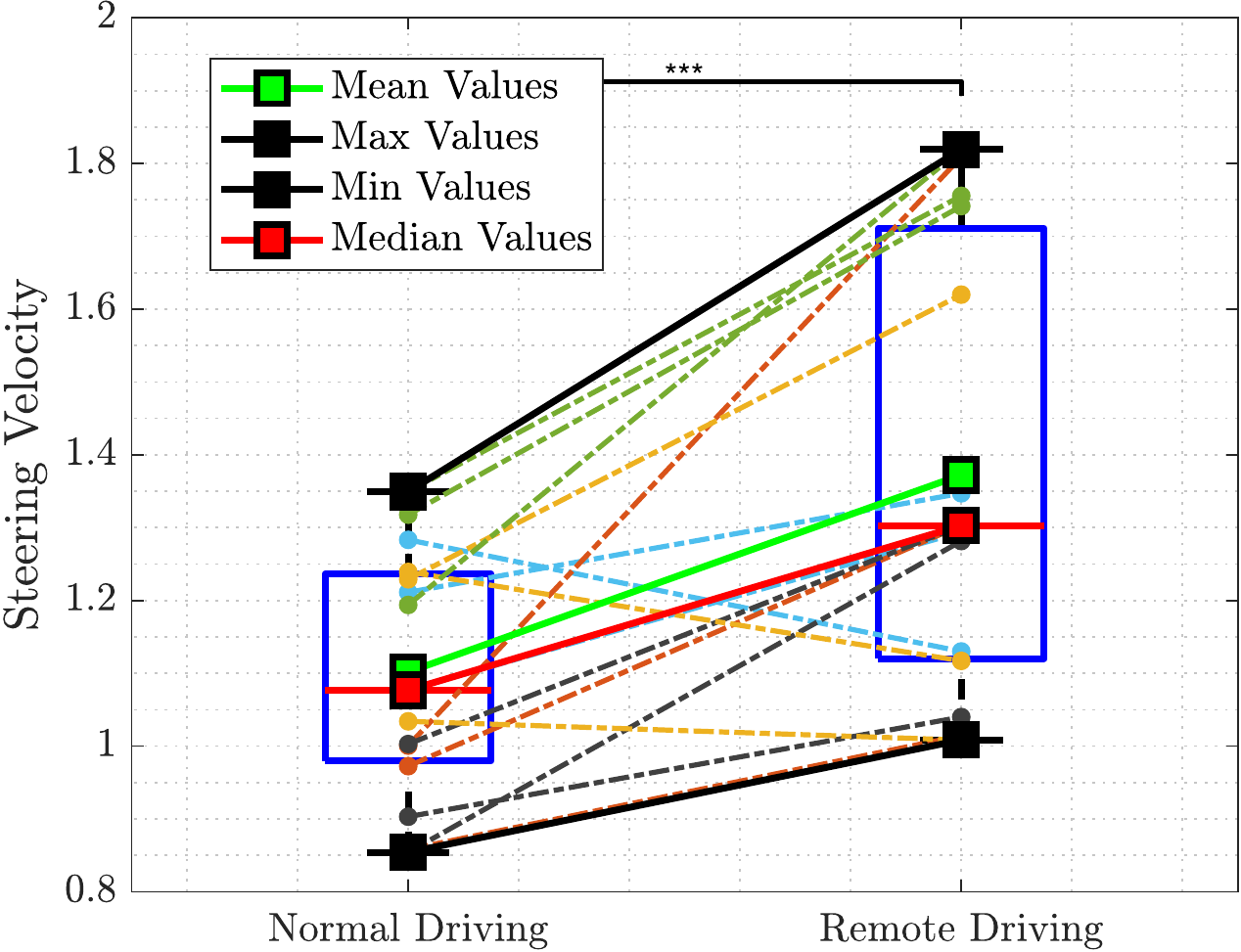}
  \caption{}
  \label{fig:SteeringVelocityComparison}
\end{subfigure}

\begin{subfigure}{0.75\linewidth}
  \centering 
  \includegraphics[width=\linewidth]{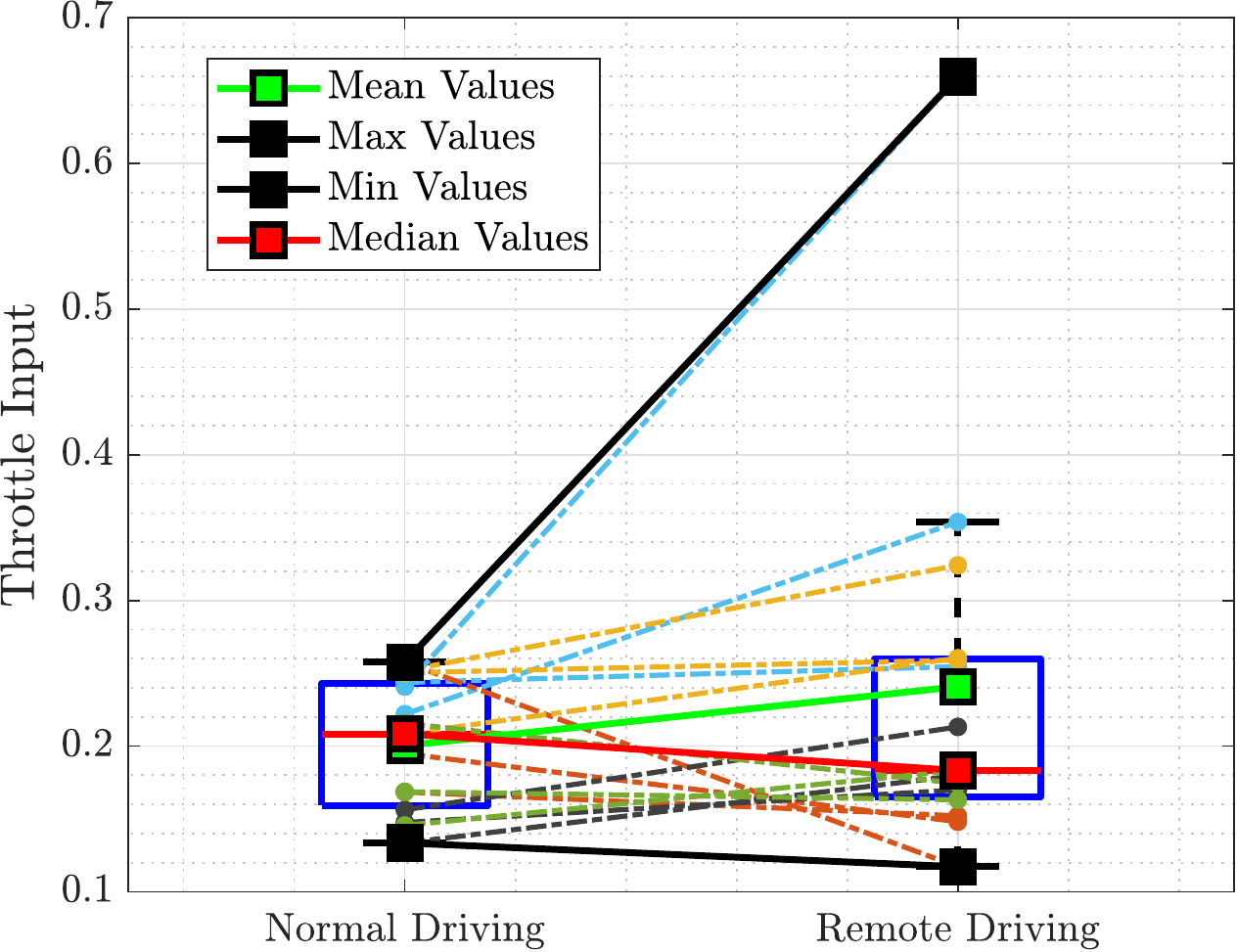}
  \caption{}
  \label{fig:ThrottleComparison}
\end{subfigure}
\caption{Comparison of (a) distance travelled, (b) steering velocity, and (c) throttle input between ND and RD. Paired sample t-test significance is presented by  $^{*}P\le0.05$,$^{**}P\le0.01$, and $^{***}P\le0.001$. [\textcolor[HTML]{77AC30}{\textbf{Driver 1}}, \textcolor[HTML]{4DBEEE}{\textbf{Driver 2}}, \textcolor[HTML]{EDB120}{\textbf{Driver 3}}, \textcolor[HTML]{D95319}{\textbf{Driver 4}}, \textcolor[HTML]{404040}{\textbf{Driver 5}}].}
\label{fig:DT_SV_DS_box}
\end{figure}

To test the hypothesis that RC, MS, SV, TI and DT are increased from normal to remote driving, a paired sample t-tested is conducted (Table \ref{tab:t_test}) to assess the significance of the results and boxplots are plotted (Figures \ref{fig:RC_MS_box} and \ref{fig:DT_SV_DS_box}). 
In the boxplots (Figures \ref{fig:RC_MS_box} and \ref{fig:DT_SV_DS_box}), the central mark in red color indicates the median, while the bottom and top edges of the box indicate the 25$^{th}$ and 75$^{th}$ percentiles, respectively. 
The whiskers extend to the most extreme data points not considered as outliers.
As mentioned before, 15 cases are studied in total (i.e., 5 different drivers testing 3 different steering feedback models) for each scenario, i.e. normal (ND) and remote (RD) driving. 
Coloured lines illustrate the effect on the metric for each case within the boxplots, where the same color refers to each driver tests (i.e., the three different steering feedback controllers). 
This aims to outline the effect consistency on the different metrics regardless of the feedback controllers, increasing the significance of the outcomes albeit the low drivers number. 


As shown in Figure \ref{fig:MotionSicknessComparison}, the values of MS in RD are significantly higher than in ND ($p_{MS}$=0.0008), which indicates that occupants would suffer more MS during RD.
This increase is consistent in most of the drivers regardless the feedback controller used, since only 1 out of the 15 cases (i.e., Driver 4 with 1 feedback controller) illustrates contradictory behaviour.
The mean MS increase is around 26 $\%$ from ND to RD.
At the same time, the RC (Figure \ref{fig:RideComfortComparison}) does not illustrate a significant difference between the two driving situations according to the paired t-test. 
Also, 4 out of 15 cases decreased ride discomfort from ND to RD. 
These cases correspond to three different drivers testing one (Driver 2 and 4) or two (Driver 3) feedback controllers.
Regarding distance travelled (DT), the increase from ND to RD is significant based on Figure \ref{fig:DistanceComparison}, indicating that in overall drivers during RD followed a wider slalom compared to ND increasing eventually the path duration. 
This could be because of the remote drivers' reduced visual awareness making it difficult to judge the distance between the physical objects (cones and car). This may cause the larger distance travelled and greater steering velocity compared to normal driving to correct the vehicle direction. The mean DT increase is around 3 $\%$ from ND to RD.
The low significance might be related with the fact that 5 out of 15 cases decreased distance travelled from ND to RD.
These five cases correspond to three different drivers testing one (Driver 2) or two (Driver 3 and Driver 5) different feedback controllers.

Regarding the objective metrics of the steering feel, the steering velocity in RD (Figure \ref{fig:SteeringVelocityComparison}), is much higher than in ND (P=0.002).
This increasing pattern is consistent regardless of the driver and the feedback controller, with only 3 out of 15 cases decreasing SV from ND to RD. 
These cases correspond to only two drivers with one (Driver 2) or two (Driver 3) feedback controllers.
The mean SV increase is around 25 $\%$ from ND to RD.
This increase is possibly caused by the driver's low situational awareness (i.e., less visual and limited motion feedback) during RD, leading to higher steering velocity and a more aggressive driving behaviour. 
This could also be the reason why the remote drivers adopted a wider slalom. 
On contrary with SV, the throttle input has no significant difference between the two scenarios (ND and RD, P=0.1384), which might be caused by various reasons. 
First of all, the intense and demanding driving required during slalom limits the drivers' decision time and they maintain the throttle at constant position.
Moreover, due to the limitations of the RCV-E, the maximum velocity was low and the drivers easily reached it. 


\begin{table}[h!]
    \centering
    \caption{Paired-sample t-test to identify if the RC, MS, SV and TI data in ND and RD come from a normal distribution with mean value equal to zero.}    
    \begin{tabular}{|c|c|c|}
         \hline
         \multicolumn{3}{|c|}{\multirow{2}{*}{\textbf{Paired-sample t-test}}} \\
         \multicolumn{3}{|c|}{}\\
         \hline 
         \textbf{Quantity}       &  \textbf{Test decision}  & \textbf{p-value} \\     
         \hline
         Ride comfort (RC)       &  Not Significant  & 0.109 \\
         \hline
         Motion sickness (MS)    &  Significant      & 0.001 \\
         \hline
         Steering velocity (SV)  &  Significant      & 0.0003 \\
         \hline
         Throttle input (TI)     &  Not significant  & 0.139 \\
         \hline
         Distance Travelled (DS) &  Significant      & 0.013 \\
         \hline         
    \end{tabular}
    \label{tab:t_test}
\end{table}

Regarding the subjective data for MS and RC and objective data for DT, SV and TI (Figures \ref{fig:RC_MS_box}-\ref{fig:DT_SV_DS_box}), on one hand, the ND gives smaller range of values, and lower median, mean, maximum and minimum values in MS (P=0.0007, Figure \ref{fig:MotionSicknessComparison}), DS (P=0.0002, Figure \ref{fig:DistanceComparison}) and SV (P=0.0002, Figure \ref{fig:SteeringVelocityComparison}) compared to the corresponding values in ND.
The similar behaviour of the three metrics indicates that the SV increase in RD can provoke more MS symptoms, while the DT is increased. 
The level of these correlations is to be validated from the more in depth analysis that will follow.
On the other hand, the changes in RC values (maximum, minimum, median and mean) (Figure \ref{fig:RideComfortComparison}) are not consistent. 
More specifically, the median, the mean and the maximum values increase, whereas the minimum value decreases in the RD compared to the ND. 
Finally, regarding the throttle input (Figure \ref{fig:ThrottleComparison}), the changes from ND to RD are the most inconsistent compared to all the studied metrics.
First of all, the range of the 25$^{th}$ and 75$^{th}$ percentiles for the throttle input is similar between the two driving scenarios. 
Meanwhile, the maximum and mean values increase from the ND to the RD scenario, whereas the median and the minimum values decrease.
This along with the high p-value obtained from the t-test makes it more difficult to extract conclusions how TI affects RC and MS.

\subsection{RC and MS correlation with steering feel objective metrics}

\begin{figure}[hp!]
\centering
\begin{subfigure}{0.75\linewidth}
  \centering 
  \includegraphics[width=\linewidth]{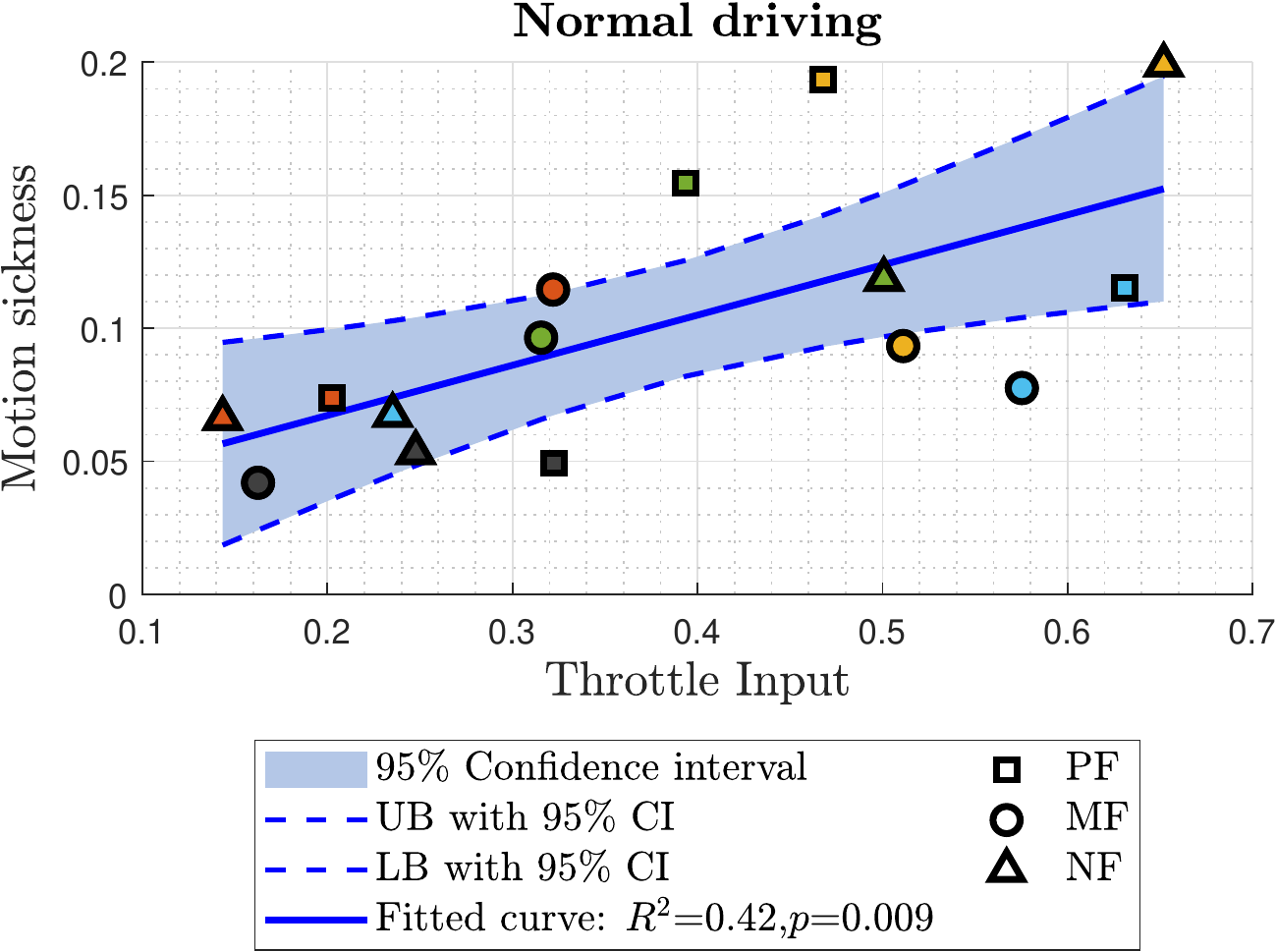}
  \caption{}
  \label{fig:ND_DT_MS_S}
\end{subfigure}
\begin{subfigure}{0.75\linewidth}
  \centering 
  \includegraphics[width=\linewidth]{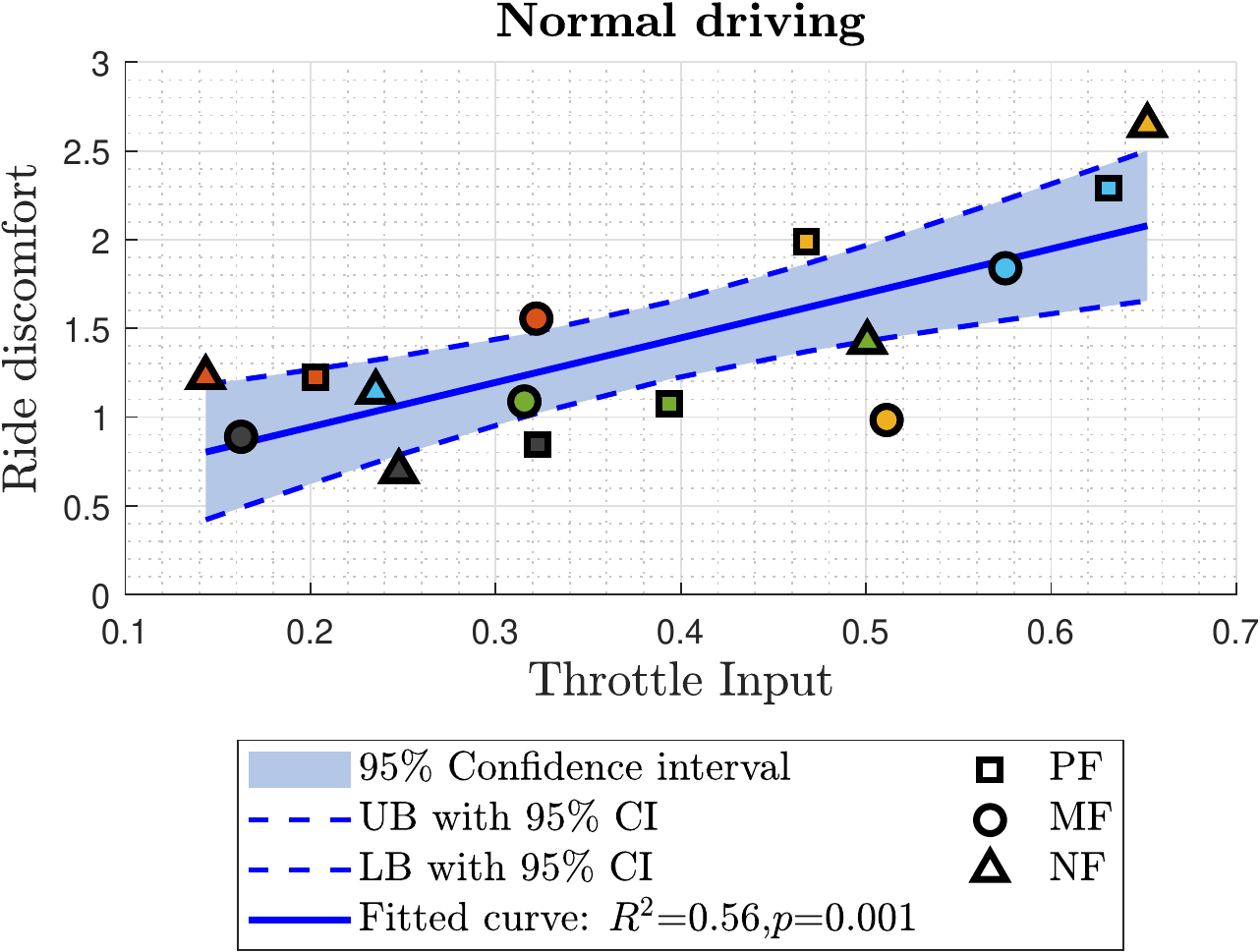}
  \caption{}
  \label{fig:ND_DT_RC_S}
\end{subfigure}
\begin{subfigure}{0.75\linewidth}
  \centering 
  \includegraphics[width=\linewidth]{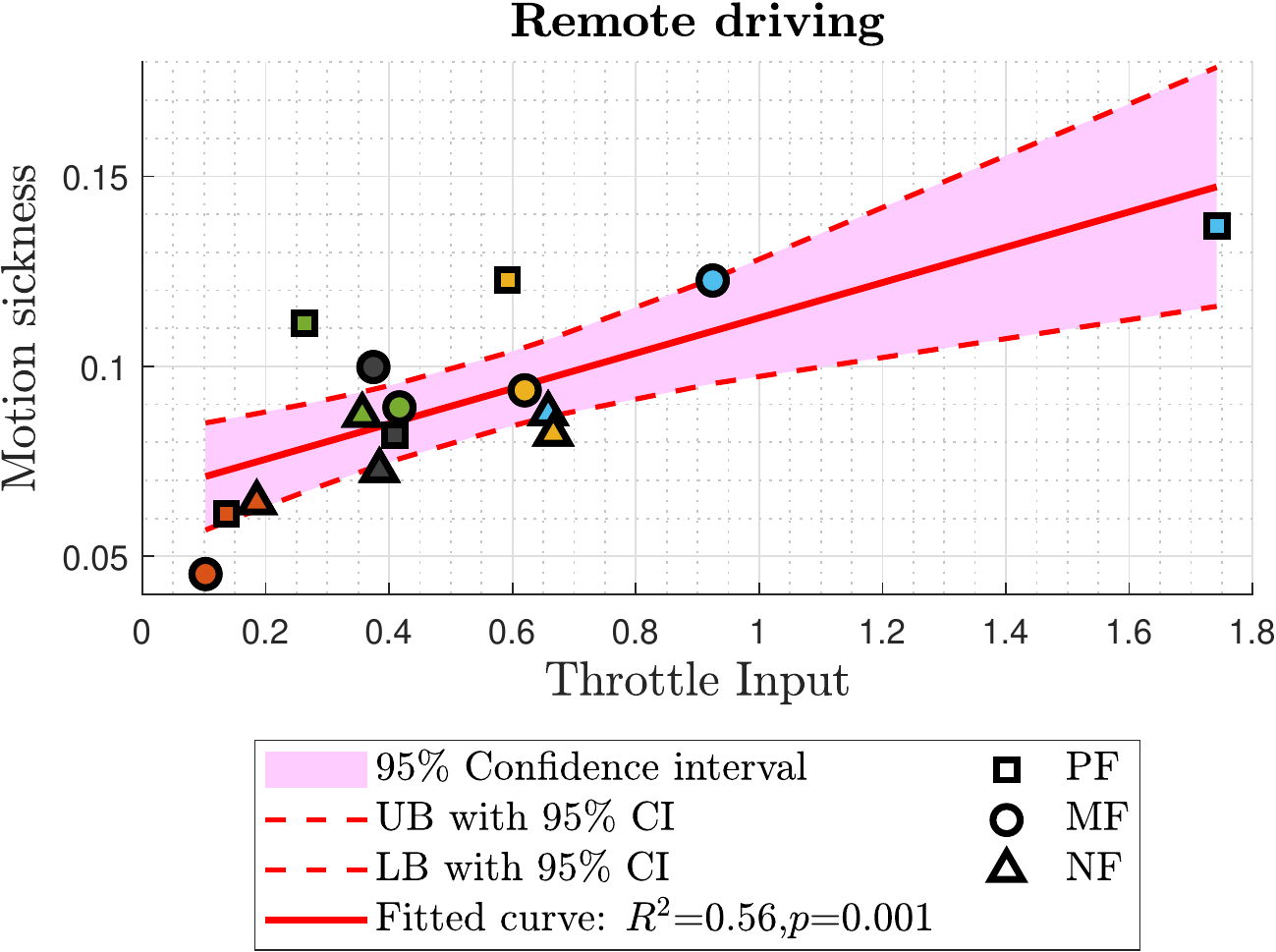}
  \caption{}
  \label{fig:RD_DT_MS_S}
\end{subfigure}
\begin{subfigure}{0.75\linewidth}
  \centering 
  \includegraphics[width=\linewidth]{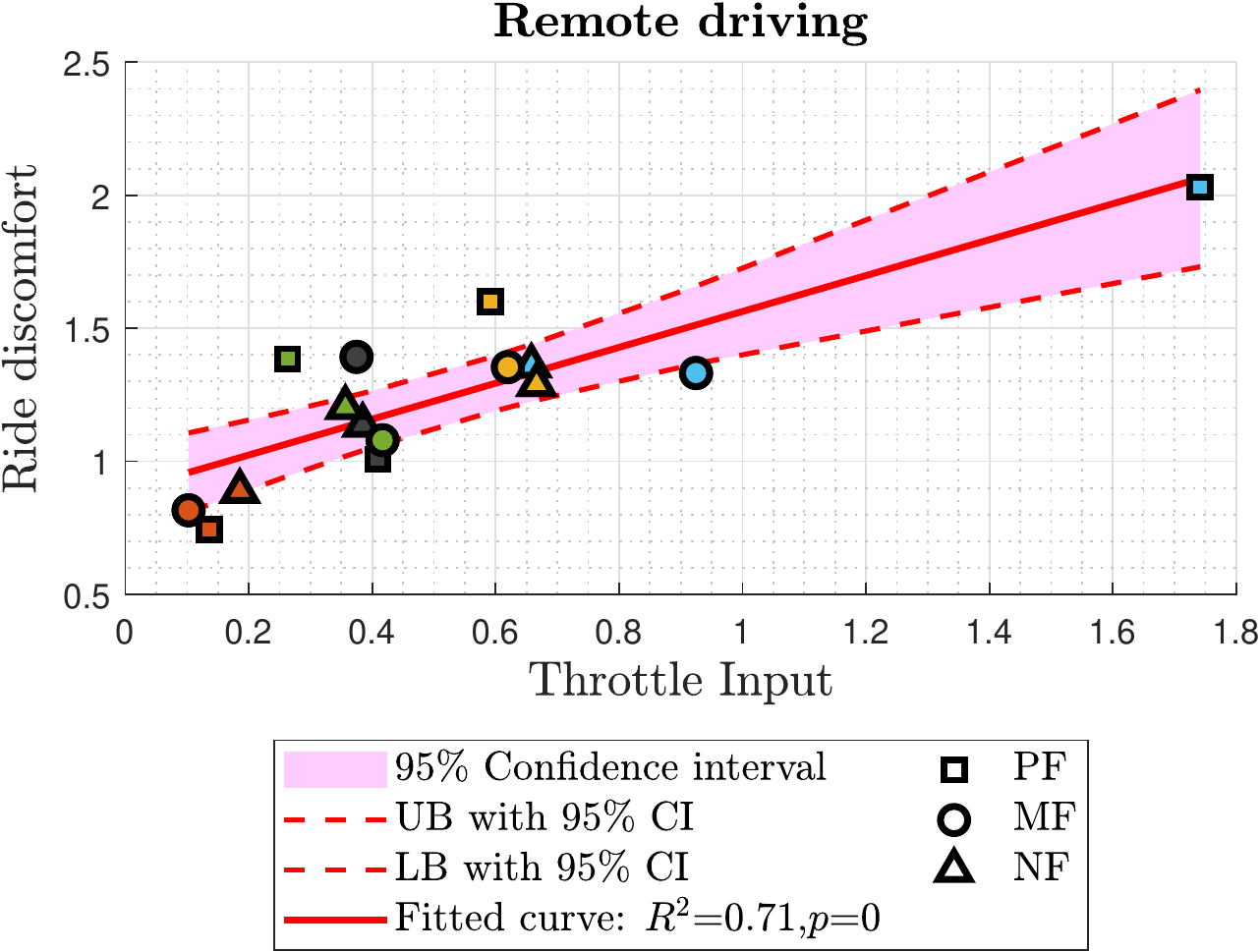}
  \caption{}
  \label{fig:RD_DT_RC_S}
\end{subfigure}
\caption{Correlation of occupants' MS with throttle input in ND (a+b) and RD (c+d) during the first 2 s, when the vehicle accelerates from standstill position. [\textcolor[HTML]{77AC30}{\textbf{Driver 1}}, \textcolor[HTML]{4DBEEE}{\textbf{Driver 2}}, \textcolor[HTML]{EDB120}{\textbf{Driver 3}}, \textcolor[HTML]{D95319}{\textbf{Driver 4}}, \textcolor[HTML]{404040}{\textbf{Driver 5}}]. \label{fig:MS_RC_DT_SV_S}}
\end{figure}

\begin{figure}[hp!]
\centering

\begin{subfigure}{0.75\linewidth}
  \centering 
  \includegraphics[width=\linewidth]{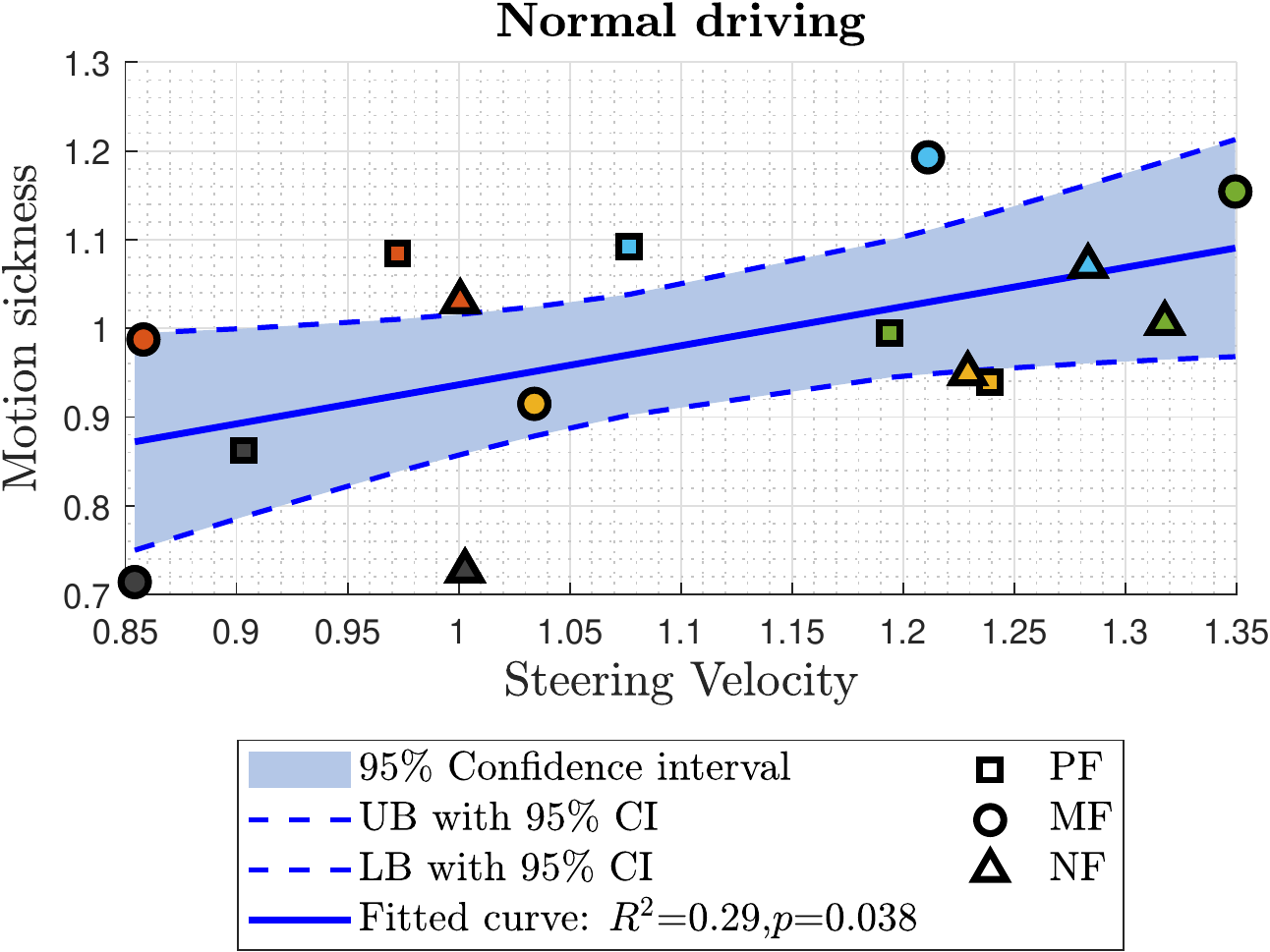}
  \caption{}
  \label{fig:ND_SV_ND_MS}
\end{subfigure}
\begin{subfigure}{0.75\linewidth}
  \centering
  \includegraphics[width=\linewidth]{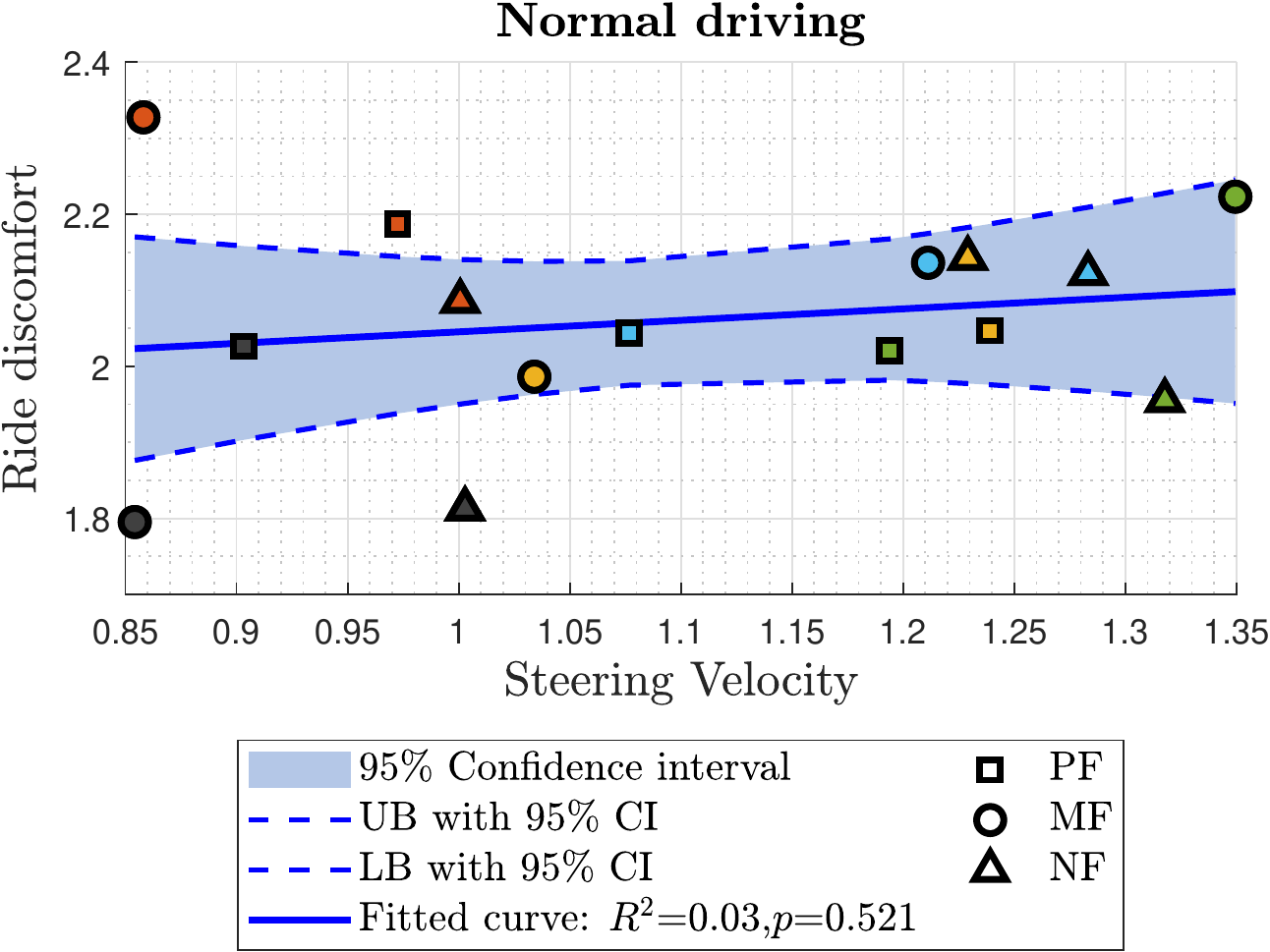}
  \caption{}
  \label{fig:ND_SV_ND_RC}
\end{subfigure}


\begin{subfigure}{0.75\linewidth}
  \centering 
  \includegraphics[width= \linewidth]{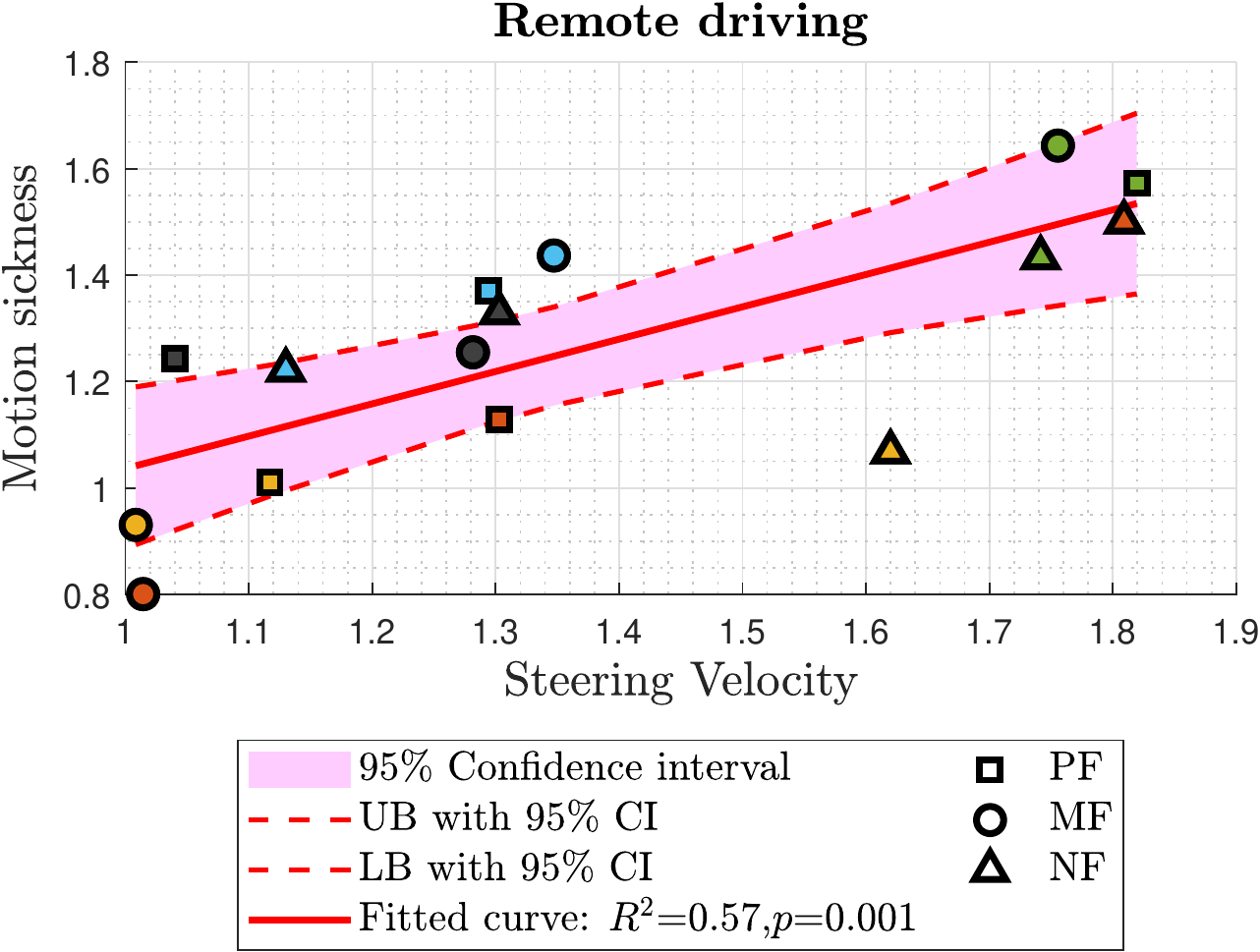}
  \caption{}
  \label{fig:RD_SV_RD_MS}
\end{subfigure}
\begin{subfigure}{0.75\linewidth}
  \centering
  \includegraphics[width= \linewidth]{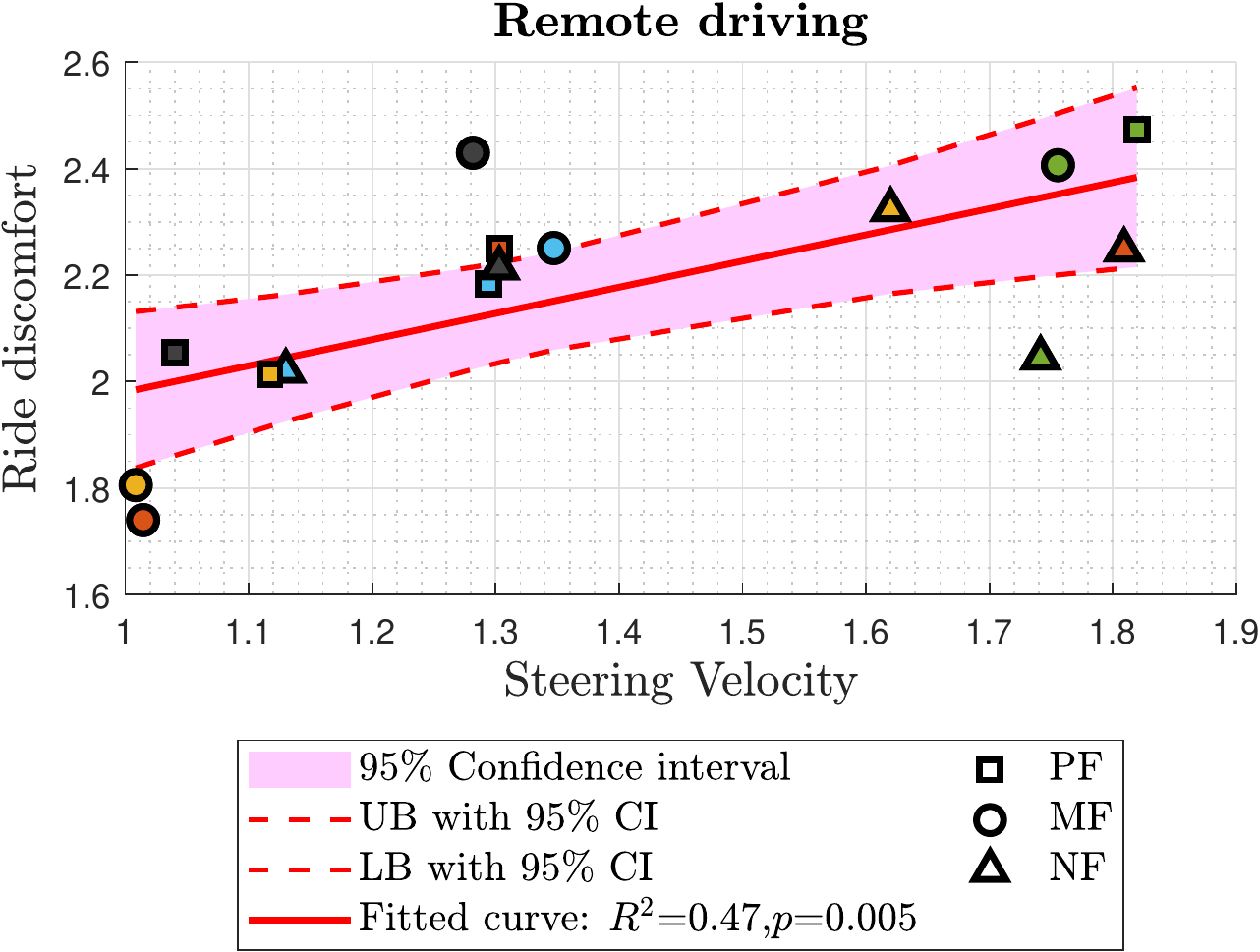}
  \caption{}
  \label{fig:RD_SV_RD_RC}
\end{subfigure}


\caption{Correlation of occupants' MS and RC with steering velocity and throttle input in ND (a-d) and RD (e-f) for the complete slalom. [\textcolor[HTML]{77AC30}{\textbf{Driver 1}}, \textcolor[HTML]{4DBEEE}{\textbf{Driver 2}}, \textcolor[HTML]{EDB120}{\textbf{Driver 3}}, \textcolor[HTML]{D95319}{\textbf{Driver 4}}, \textcolor[HTML]{404040}{\textbf{Driver 5}}]. \label{fig:MS_RC_DT_SV}}
\end{figure}

In this section, the relation of RC and MS with SV and TI is investigated using linear regression analysis in time spans of the ride: (a) 0 - 2 s, where the vehicle accelerates from standstill (Figure \ref{fig:MS_RC_DT_SV_S}), and (b) the complete slalom (Figure \ref{fig:MS_RC_DT_SV}). 
For the latter, the remarks are validated using stepwise regression (Table \ref{tab:multiple_regression1}).
Albeit not emphasising the impact of the different feedback controllers (PF, MF and NF) on RC and MS, the data used (fifteen cases, five drivers testing three steering feedback controllers) for the linear regression analysis are plotted in Figure \ref{fig:MS_RC_DT_SV_S}-\ref{fig:MS_RC_DT_SV} with different marker face color per driver and different markers style per feedback controller. 
For example, the three yellow markers (one square, one circle and one triangle) refer to Driver 3 performance for the three different steering feedback controllers (PF, MF and NF), while the five triangle markers (green, blue, yellow, orange and grey) illustrate the NF case for all five drivers. 
Overall, the different cases (i.e., different driver with different controllers) are mostly consistent regarding their effect on the various objective metrics studied when shifting from normal to remote driving regardless the driver or the feedback controller.
This consistency secures the significance of the outcome despite the low number of drivers. 

Regarding the first part, where the vehicle accelerates from standstill, the emphasis is on the correlation of MS and RC with the throttle input variations (TI) since there are no variations in steering during this time period. 
According to Figure \ref{fig:MS_RC_DT_SV_S}, MS and RC are highly correlated with TI with also great significance (R$^2$ = 0.42/ P=0.009, Figure \ref{fig:ND_DT_MS_S} and R$^2$ = 0.56/ P=0.001, Figure \ref{fig:ND_DT_RC_S}). 
The correlation is larger when it comes to RC.
Furthermore, the drivers' throttle behaviour in the different cases is very scattered, which can be due to the fact that there was more situational awareness or vehicle motion feedback to the drivers. 
Meanwhile, the occupants' ride comfort and motion sickness are correlated more in RD (R$^2$ = 0.56/ P=0.001, Figure \ref{fig:RD_DT_MS_S} and R$^2$ = 0.71/ P=0.0001, Figure \ref{fig:RD_DT_RC_S}).
The high correlation of MS and RC with TI with great significance is consistent (R$^2$ = 0.36/ P=0.03 and R$^2$ = 0.39/ P=0.02, Figure \ref{fig:RD_DT_RC_S}) if calculated in the [10 90] percentile of the data where the outlier (Driver 2 with PF with TI = $\sim$ 1.8) is removed. 
At the same time, the driver's behaviour does not differ significantly as expected, since nothing differentiated that could affect the remote drivers' throttle behaviour between the cases. 
The similar behaviour could also be affected by the fact that the drivers' do not receive any feedback about the vehicle motion that could help them adjust their throttle input, except from microphone sound. 

Regarding the complete slalom, the focus is shifted on the correlation of MS and RC with the steering velocity (SV) since there are insignificant variations in the throttle input after the first seconds (acceleration from standstill), when the maximum velocity (18 km/h) is reached. 
This was outlined in the correlations explored between MS and RC with TI in both ND and RD. 
More specifically, MS is insignificantly correlated with TI in ND (R$^2$ = 0.14, / P=0.162), while in RD there is no correlation at all (R$^2$ = 0/ P=0.876). 
Similarly, RC illustrates decent correlation with TI (R$^2$ = 0.21/ P=0.083) in ND, but no correlation in RD (R$^2$ = 0/ P=0.985).
The correlation plots of RC and MS with TI in ND and RD are not included for the shake of simplicity, and the emphasis is on their correlation with SV. 
According to Figure \ref{fig:MS_RC_DT_SV}, MS is marginally correlated with SV in both ND and RD, (R$^2$ = 0.29/ P=0.058, Figure \ref{fig:ND_SV_ND_MS} and R$^2$ = 0.57/ P=0.162, Figure \ref{fig:RD_SV_RD_MS}).
Additionally, the LB and UB fitted curves (i.e., the fitted curves with lower and upper coefficient bounds with 95 $\%$ confidence level) display small distances from the main curve, illustrating high confidence in the result. 
The trend of these curves is the same with the one of the main curve, illustrating consistency in the increasing pattern of MS and SV. 
Regarding the RC correlation with SV, the results are inconsistent compared to the MS correlation with the same metrics. 
RC presents no correlation with SV (R$^2$ = 0.03/ P=0.521, Figure \ref{fig:ND_SV_ND_RC}), but the opposite behaviour is identified in RD, where the RC metric is correlated highly with SV but not at all with TI (R$^2$ = 0.47/ P=0.006, Figure \ref{fig:RD_SV_RD_RC}).
In addition to the decent correlation of RC with SV, the trend of the fitted curve with the one of the LB and UB curves is consistent. 
Also, the distances of the LB and UB curves with the main are small, indicating high prediction probability.

To sum up, based on the above, both the steering and throttle behaviour affect the incidence of MS and RC in ND.
Meanwhile, MS and RC correlation with SV is increased from ND to RD, whereas their correlation with TI is decreased.
On the other hand, in RD, SV is the main factor to be correlated with RC and MS in greater levels than the ones in ND, with the throttle variations not affecting the overall motion comfort.  
To validate this, a multiple regression model (Table \ref{tab:multiple_regression1}) is developed to investigate MS and RC relation with SV and TI during ND and RD. 
In ND, RC and MS are predicted based on the interaction of SV with TI (SV*TI) rather than the combination of both as separate terms. 
Whereas in RD, albeit the model is able to identify interactions both RC and MS depend exclusively on SV. 
The above remarks are potentially related with the difference in the RD behaviour. 
The remote driver's limited view in RD led to higher SV, provoking higher MS (as illustrated in Figure \ref{fig:SteeringVelocityComparison} and \ref{fig:MotionSicknessComparison}) and leading to greater correlation levels compared to ND. 
As mentioned before, the remote drivers' limited view and enviromental awareness, makes it difficult for them to judge the distance to physical objects, causing greater deviation that lead to larger steering velocity inputs to correct the vehicle direction.
On the other hand, due to the remote drivers' limited situation awareness and the lack of feedback about the vehicle velocity, the throttle input varied inconsistently compared to the ND.
This contributed less to the MS accumulation decreasing the TI correlation levels with MS.
This could also be affected by other factors such as: (a) the small decision time for the drivers during the slalom, which makes them to maintain the throttle at constant position, and (b) the maximum allowed velocity of the RCV-E. 
An experiment having more intense longitudinal effects could provide more concrete conclusions.

\begin{table}[h!]
    \centering
    \caption{Multiple regression of RC and MS with SV.}    
    \begin{tabular}{|c|c|c|}
         \hline
         \multicolumn{3}{|c|}{\multirow{2}{*}{\textbf{Normal Driving}}} \\
         \multicolumn{3}{|c|}{}\\
         \hline
         \textbf{RC}        &  y $\sim$ 1 + SV*TI      & R$^2$ = 0.58, p = 0.0195 \\
         \hline
         \textbf{MS}        &  y $\sim$ 1 + SV*TI      & R$^2$ = 0.54, p = 0.0326 \\         
         \hline
         \multicolumn{3}{|c|}{\multirow{2}{*}{\textbf{Remote Driving}}} \\
         \multicolumn{3}{|c|}{}\\         
         \hline
         \textbf{RC}        &  y $\sim$ 1 + SV         & R$^2$ = 0.47, p = 0.0051 \\
         \hline
         \textbf{MS}        &  y $\sim$ 1 + SV         & R$^2$ = 0.57, p = 0.0012 \\
         \hline
    \end{tabular}
    \label{tab:multiple_regression1}
\end{table}

\subsection{Correlation between subjective steering feel (SA) and objective metrics for motion comfort (RC and MS)}


\renewcommand\tabcolsep{12.0pt}
\begin{table*}[h!]
\centering
\caption {R$^2$ and \textit{p}-value for significance of the correlations between subjective driver feel assessment (Table \ref{tab:SA_Metrics}) and RC/MS metrics. The coloured cells have correlated the metrics within the [10 90] percentile of the data removing outliers \cite{Liao2016}. \label{tab:RSquareOfCorrelation}}
\begin{center}
\begin{tabular}{ccllllllll}
\toprule
               &                 & \multicolumn{4}{c}{\textbf{ND}}                                                                                                           & \multicolumn{4}{c}{\textbf{RD}}                                                                                                           \\ \cmidrule{3-10} 
               &                 & \multicolumn{2}{c}{\textbf{RC}}                                     & \multicolumn{2}{c}{\textbf{MS}}                                     & \multicolumn{2}{c}{\textbf{RC}}                                     & \multicolumn{2}{c}{\textbf{MS}}                                     \\ \cmidrule{3-10} 
\textbf{Level} & \textbf{Metric} & \multicolumn{1}{c}{\textbf{R$^2$}} & \multicolumn{1}{c}{\textbf{\textit{p}}} & \multicolumn{1}{c}{\textbf{R$^2$}} & \multicolumn{1}{c}{\textbf{\textit{p}}} & \multicolumn{1}{c}{\textbf{R$^2$}} & \multicolumn{1}{c}{\textbf{\textit{p}}} & \multicolumn{1}{c}{\textbf{R$^2$}} & \multicolumn{1}{c}{\textbf{\textit{p}}} \\ \midrule
\textbf{1}     & \textbf{SA01}   & 0.20                               & 0.097                          & 0.23                               & 0.068                          & 0.16                               & 0.370                          & 0.03                               & 0.510                          \\ \midrule
\textbf{2}     & \textbf{SA10}   & \cellcolor[HTML]{E7E6E6}0.15       & \cellcolor[HTML]{E7E6E6}0.191 & \cellcolor[HTML]{E7E6E6}0.33       & \cellcolor[HTML]{E7E6E6}0.039  & \cellcolor[HTML]{E7E6E6}0.11       & \cellcolor[HTML]{E7E6E6}0.275  & \cellcolor[HTML]{E7E6E6}0.05       & \cellcolor[HTML]{E7E6E6}0.478  \\ \midrule
\textbf{3}     & \textbf{SA11}   & \cellcolor[HTML]{E7E6E6}0.25       & \cellcolor[HTML]{E7E6E6}0.083 & \cellcolor[HTML]{E7E6E6}0.25       & \cellcolor[HTML]{E7E6E6}0.083  & 0.26                               & 0.053                          & 0.03                               & 0.512                          \\ \midrule
\textbf{3}     & \textbf{SA12}   & \cellcolor[HTML]{E7E6E6}0.27       & \cellcolor[HTML]{E7E6E6}0.067 & \cellcolor[HTML]{E7E6E6}0.39       & \cellcolor[HTML]{E7E6E6}0.023  & \cellcolor[HTML]{E7E6E6}0.24       & \cellcolor[HTML]{E7E6E6}0.086  & \cellcolor[HTML]{E7E6E6}0.02       & \cellcolor[HTML]{E7E6E6}0.666  \\ \midrule
\textbf{2}     & \textbf{SA20}   & 0.29                               & 0.039                          & 0.32                               & 0.029                          & 0.17                               & 0.121                          & 0.07                               & 0.346                          \\ \midrule
\textbf{3}     & \textbf{SA21}   & 0.32                               & 0.029                          & 0.18                               & 0.110                          & 0.00                               & 0.975                          & 0.00                               & 0.987                          \\ \midrule
\textbf{3}     & \textbf{SA22}   & 0.09                               & 0.292                          & 0.11                               & 0.226                          & 0.07                               & 0.333                          & 0.04                               & 0.461                          \\ \midrule
\textbf{2}     & \textbf{SA30}   & 0.07                               & 0.346                          & 0.06                               & 0.370                          & \cellcolor[HTML]{E7E6E6}0.13       & \cellcolor[HTML]{E7E6E6}0.234  & \cellcolor[HTML]{E7E6E6}0.02       & \cellcolor[HTML]{E7E6E6}0.679 \\ \midrule
\textbf{3}     & \textbf{SA31}   & \cellcolor[HTML]{E7E6E6}0.26      & \cellcolor[HTML]{E7E6E6}0.073  & \cellcolor[HTML]{E7E6E6}0.61      & \cellcolor[HTML]{E7E6E6}0.002  & \cellcolor[HTML]{E7E6E6}0.00      & \cellcolor[HTML]{E7E6E6}0.861  & \cellcolor[HTML]{E7E6E6}0.33      & \cellcolor[HTML]{E7E6E6}0.042  \\ \midrule
\textbf{3}     & \textbf{SA32}   & \cellcolor[HTML]{E7E6E6}0.28      & \cellcolor[HTML]{E7E6E6}0.061  & \cellcolor[HTML]{E7E6E6}0.68      & \cellcolor[HTML]{E7E6E6}0.001  & \cellcolor[HTML]{E7E6E6}0.10      & \cellcolor[HTML]{E7E6E6}0.292  & \cellcolor[HTML]{E7E6E6}0.24      & \cellcolor[HTML]{E7E6E6}0.090  \\ 
\bottomrule
\end{tabular}
\end{center}
\end{table*} 

\begin{figure}[h!]
\centering
\begin{subfigure}{0.65\linewidth}
  \centering 
  \includegraphics[width=\linewidth]{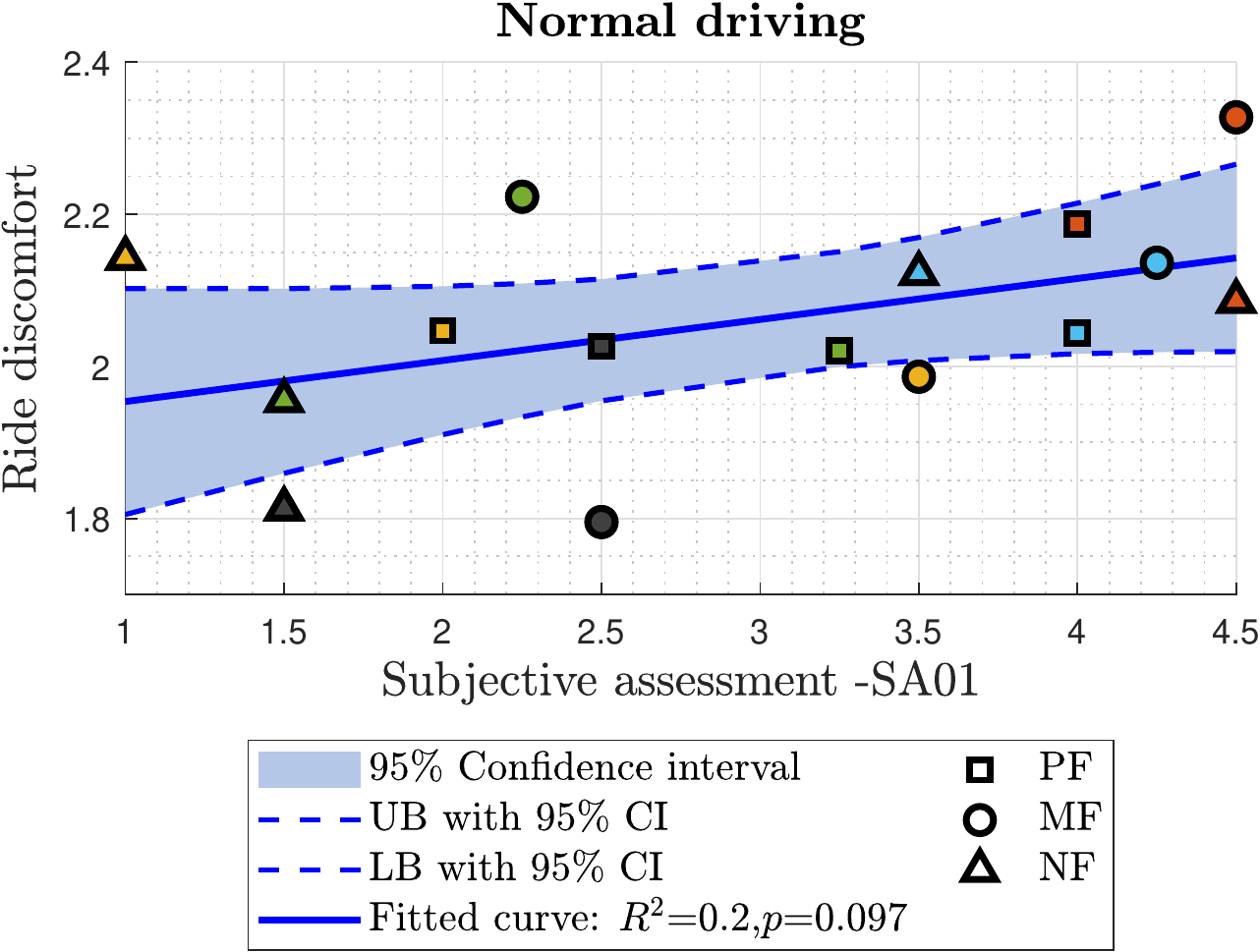}
  \caption{}
  \label{fig:ND_SA_ND_MS_SA01_S}
\end{subfigure}
\begin{subfigure}{0.65\linewidth}
  \centering 
  \includegraphics[width=\linewidth]{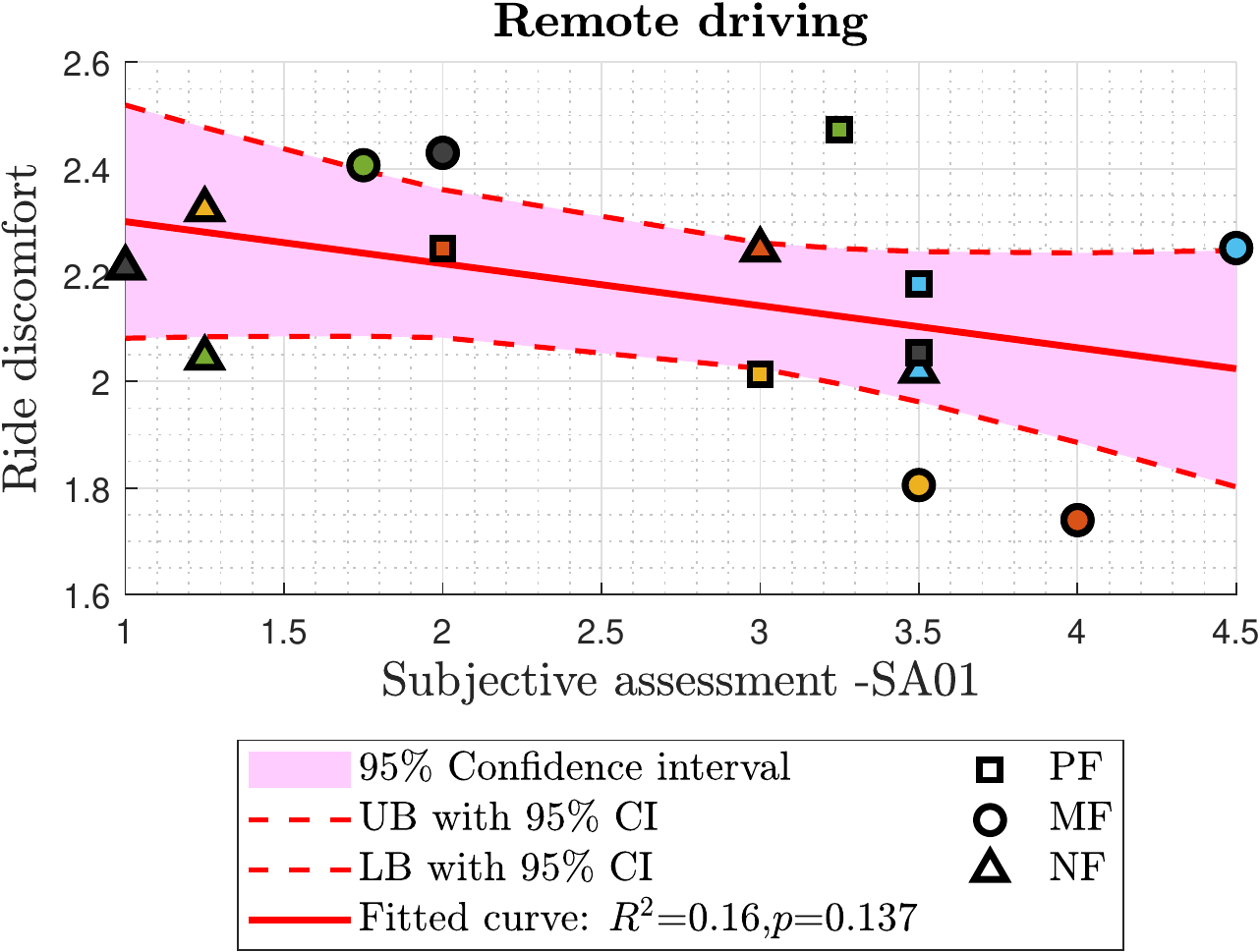}
  \caption{}
  \label{fig:RD_SA_RD_MS_SA01_S}
\end{subfigure}
\caption{Correlation of the overall subjective steering feel (SA01) with occupants' RC in (a) ND and (b) RD. The fitting is on the data belonging to [0 100] percentile.}\label{fig:RC_MS_SA01_S}
\end{figure}
\begin{figure}[h!]
\centering
\begin{subfigure}{0.65\linewidth}
  \centering 
  \includegraphics[width=\linewidth]{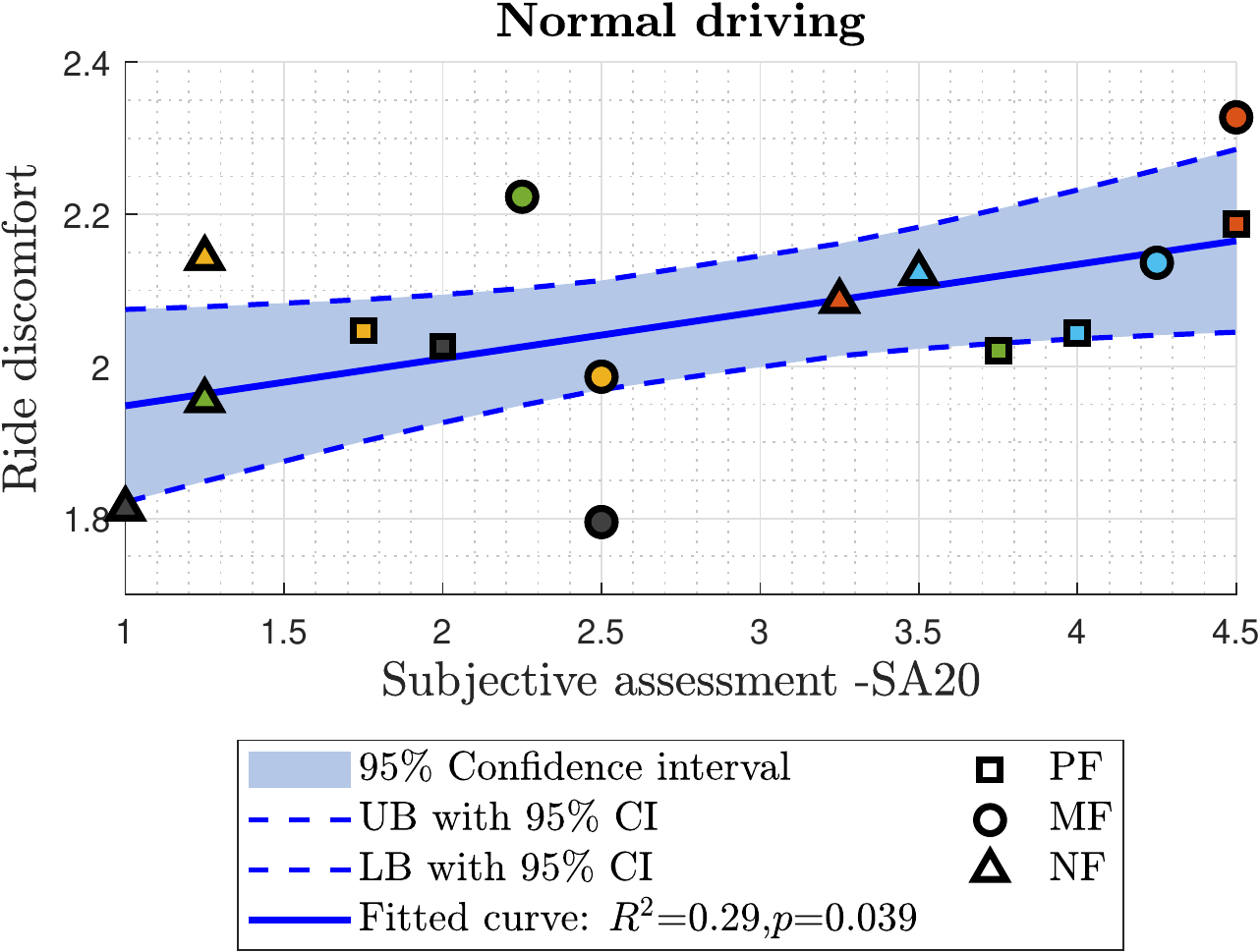}
  \caption{}
  \label{fig:ND_SA_ND_RC_SA20_S}
\end{subfigure}
\begin{subfigure}{0.65\linewidth}
  \centering 
  \includegraphics[width=\linewidth]{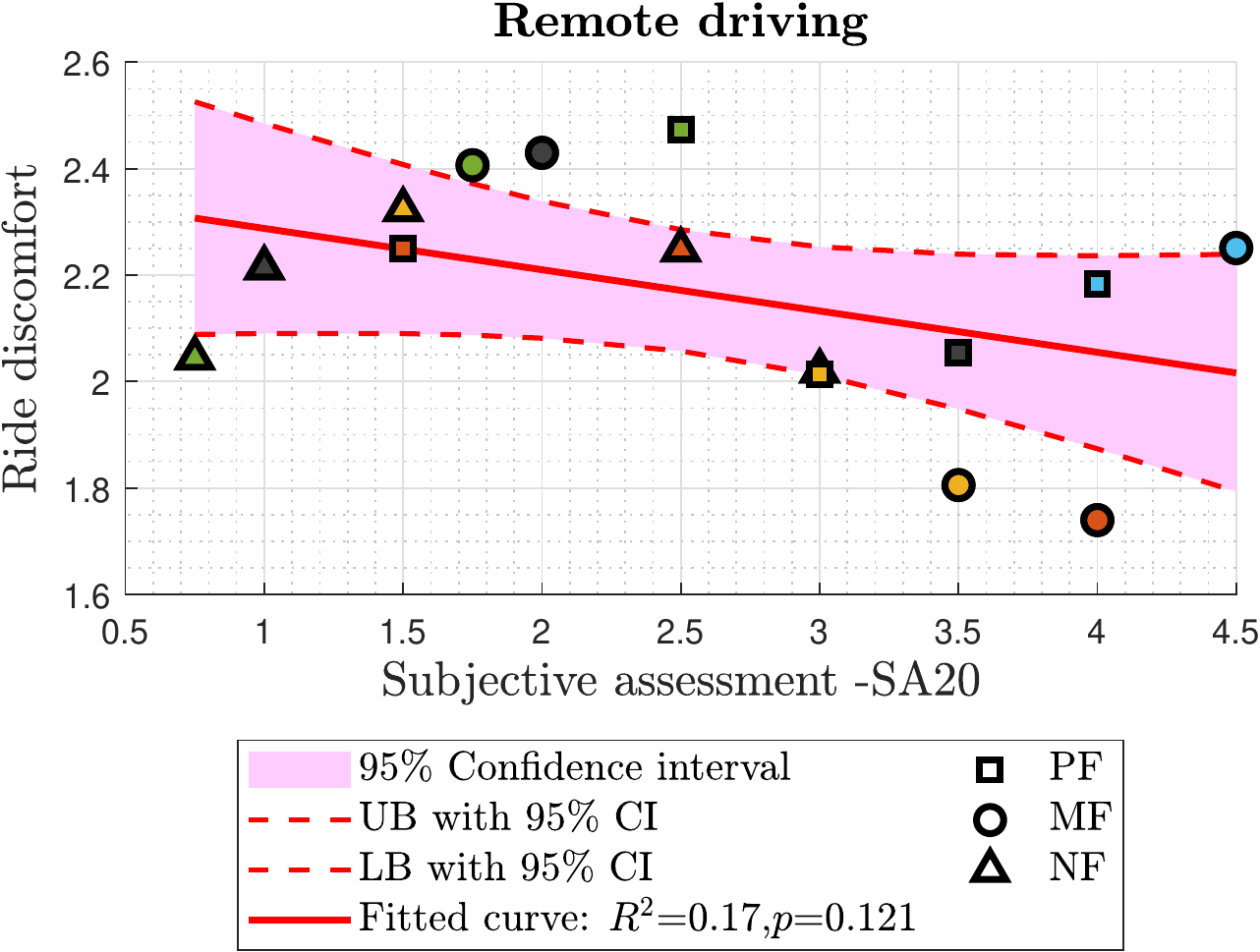}
  \caption{}
  \label{fig:RD_SA_RD_RC_SA20_S}
\end{subfigure}
\caption{Correlation of occupants ride comfort with subjective level of steering wheel characteristic feel (SA20) in (a) ND and  (b) RD. The fitting is on the data belonging  to [0 100] percentiles. \label{fig:RC_MS_SA20_S}}
\end{figure}
\begin{figure}[h!]
\centering
\begin{subfigure}{0.65\linewidth}
  \centering 
  \includegraphics[width=\linewidth]{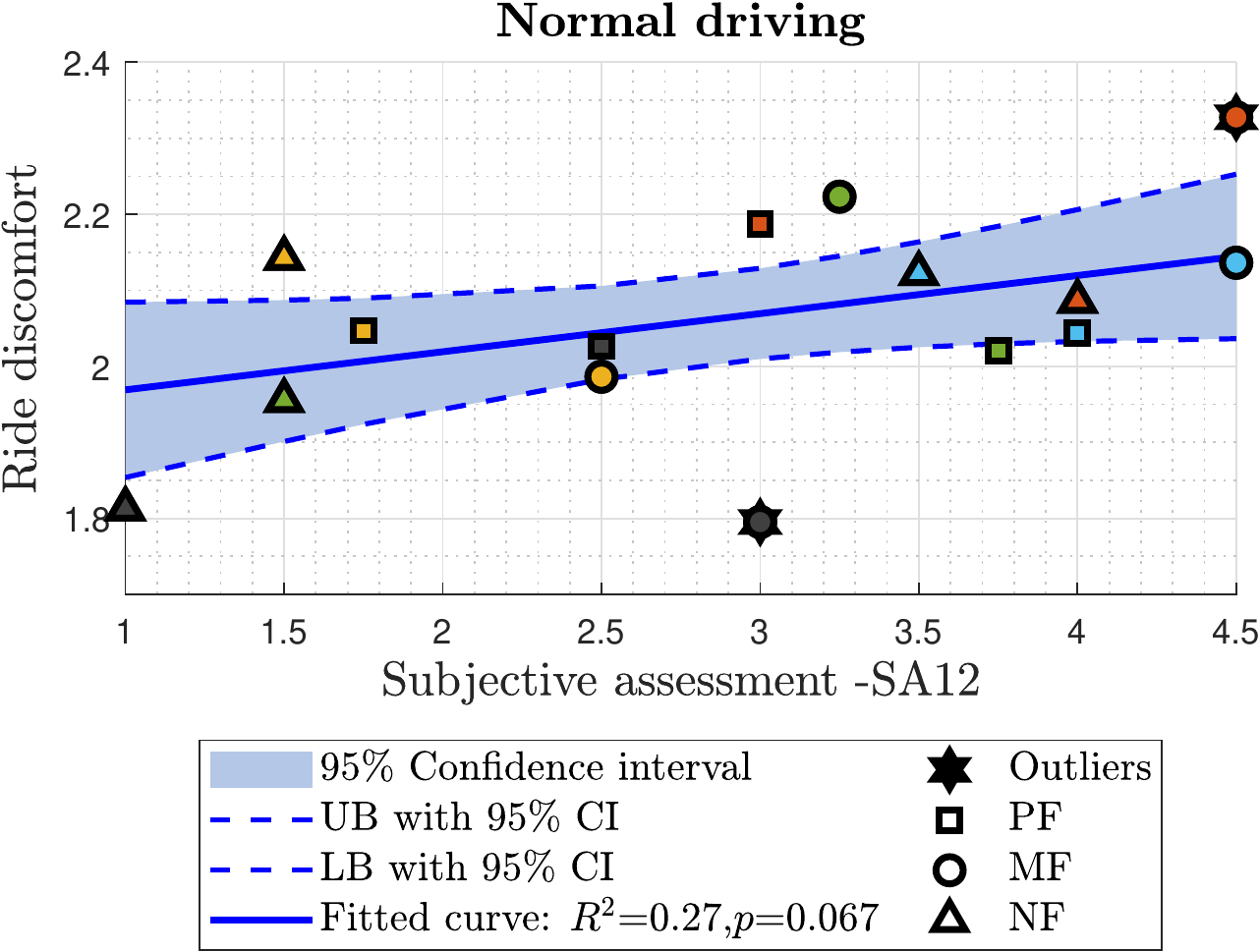}
  \caption{}
  \label{fig:ND_SA_ND_RC_SA12_S}
\end{subfigure}
\begin{subfigure}{0.65\linewidth}
  \centering 
  \includegraphics[width=\linewidth]{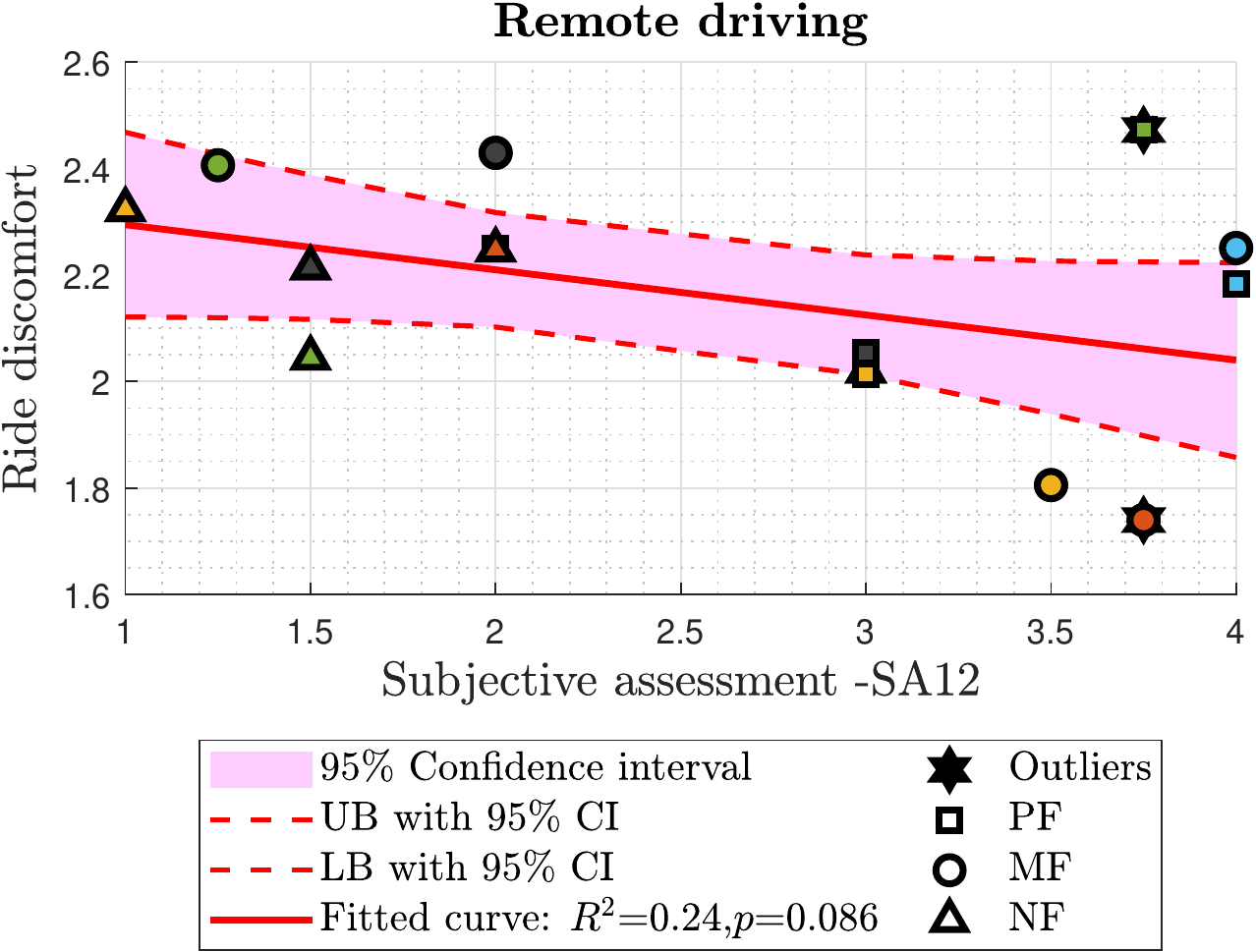}
  \caption{}
  \label{fig:RD_SA_RD_RC_SA12_S}
\end{subfigure}
\caption{Correlation of occupants' ride comfort with the subjective level of steering feedback communication (SA12) in (a) ND and (b) RD. The fitting is on the data belonging to [10 90] percentile. \label{fig:RC_MS_SA12_S}}
\end{figure}
\begin{figure}[h!]
\centering
\begin{subfigure}{0.65\linewidth}
  \centering 
  \includegraphics[width=\linewidth]{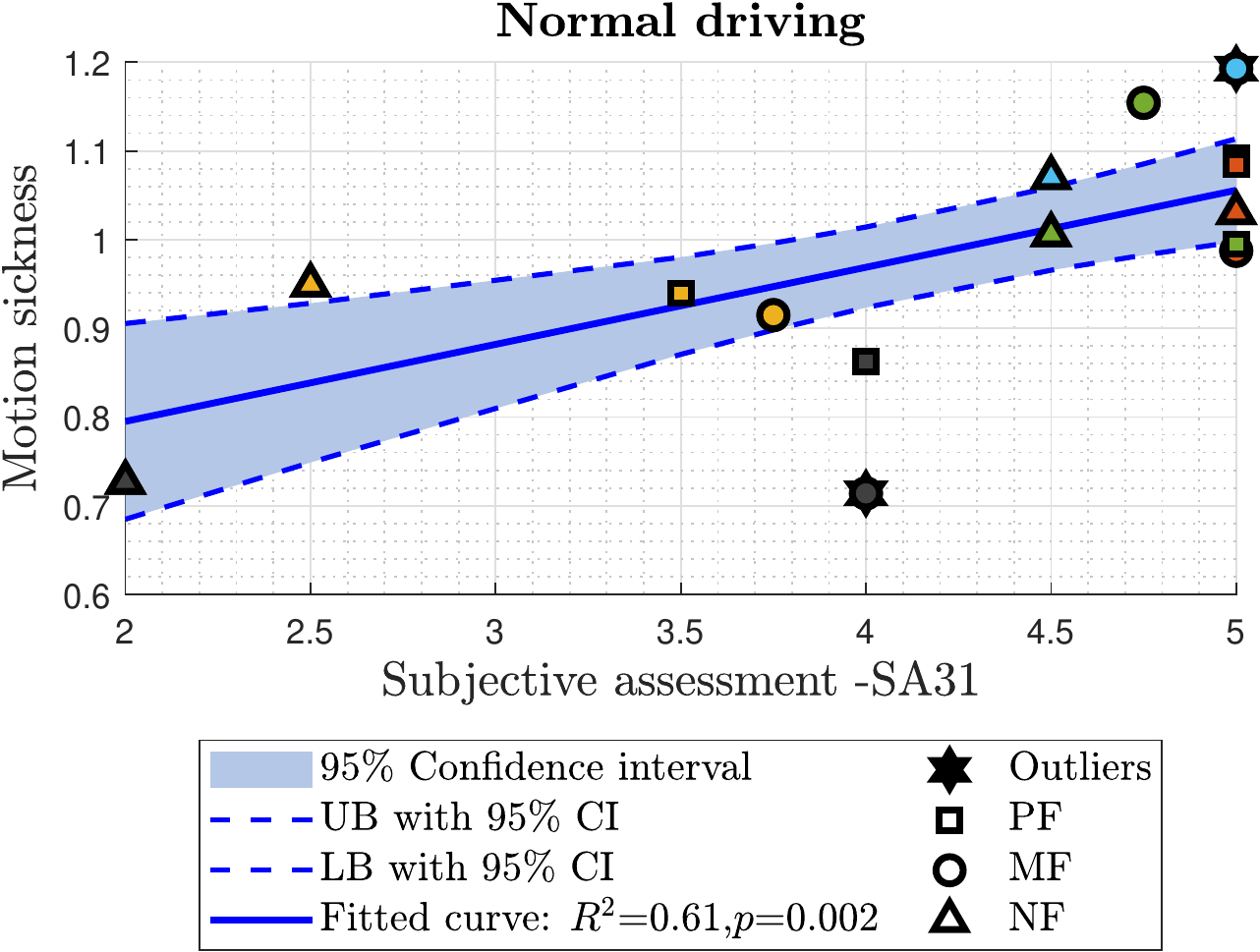}
  \caption{}
  \label{fig:ND_SA_ND_MS_SA31_S}
\end{subfigure}
\begin{subfigure}{0.65\linewidth}
  \centering 
  \includegraphics[width=\linewidth]{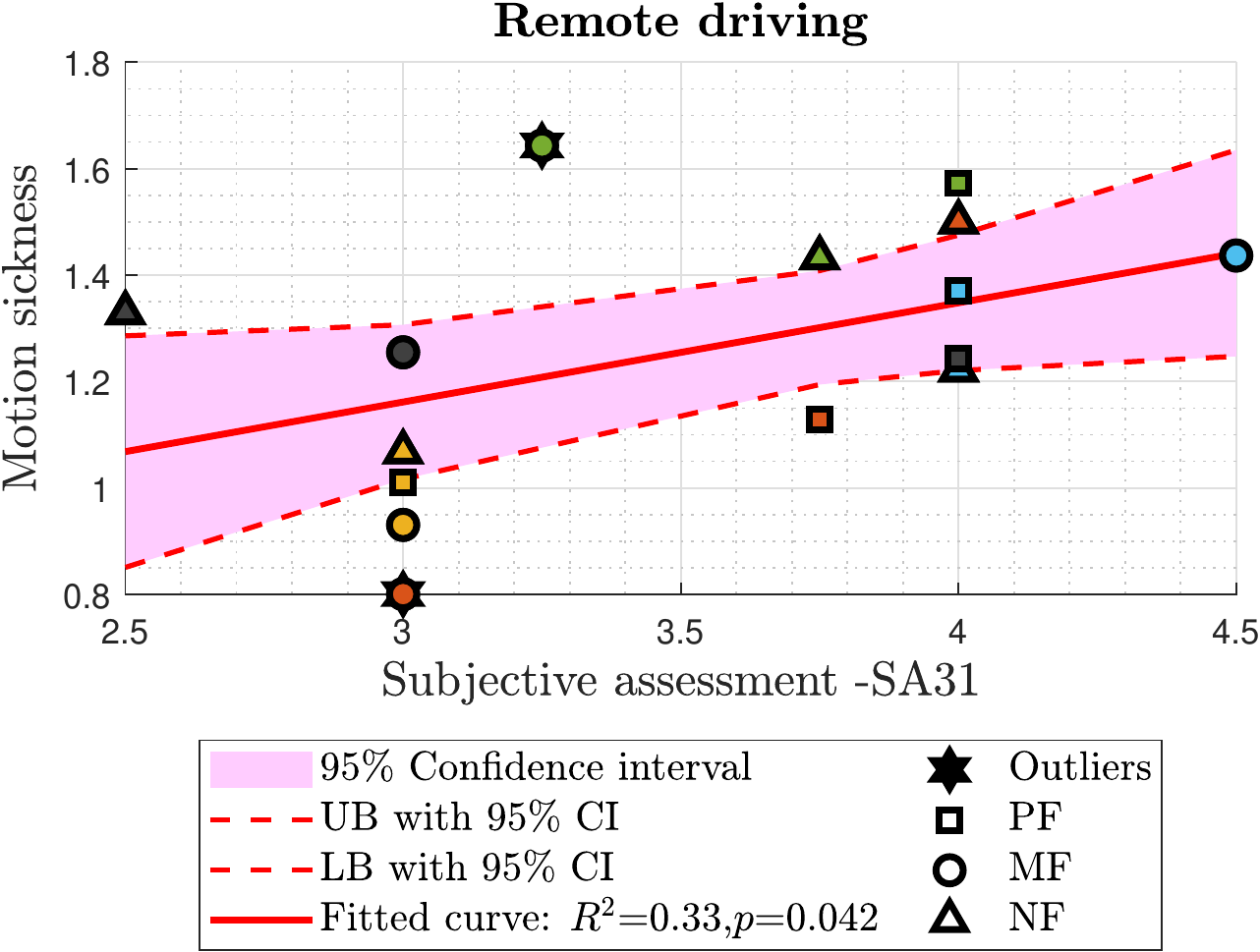}
  \caption{}
  \label{fig:RD_SA_RD_MS_SA31_S}
\end{subfigure}
\caption{Correlation of occupants sickness with subjective assessment of the degree of success accomplishing the task (SA31) in (a) ND and (b) RD. The fitting is on the data belonging within [10 90] percentile. \label{fig:RC_MS_SA31_S}}
\end{figure}

This section focuses on SA metrics correlation with RC and MS through linear regression analysis to unravel their relation.
For this, firstly, Table \ref{tab:RSquareOfCorrelation} illustrates the regression levels using R$^2$ for all SA metrics with RC and MS in ND and RD. 
Then, the fitted curves of the ones with higher R$^2$ are presented (Figures \ref{fig:RC_MS_SA01_S}-\ref{fig:RC_MS_SA31_S}), and they are also plotted for lower and upper coefficient bounds with 95 $\%$ confidence level.
Similarly with before, the data used (fifteen cases, five drivers testing three steering feedback controllers) for the linear regression analysis are plotted with different marker face color per driver and different markers style per feedback controller. 

According to Table \ref{tab:RSquareOfCorrelation}, the SA metrics correlation with RC and MS varies significantly between ND and RD. 
The R$^2$ is within low and medium levels in both scenarios (ND and RD), while the subjective driver feel assessment (SA metrics) is correlated with MS and RC mostly in ND than RD (R$_{ND}^2>$R$_{RD}^2$). 
In ND, SA metrics are more correlated with MS rather than RC, which might be because the driving behaviour is proven to affect more the low frequencies that are related to MS \cite{Htike2022}.
At the same time, there is principally no correlation of SA metrics with MS in RD, except from SA31 and SA32 both related with drivers' confidence and control.
On the other hand, RC illustrates some correlation with SA metrics in RD, but the levels are lower compared to ND.
More specifically, despite the low correlation of RC, RC correlates amply (R$^2$=~0.25) with the Level 3 safety assessment metrics (SA11 and SA12) in RD similarly with ND. 
At the same time, there insignificant or no correlation with other SA metrics about confidence and control, or steering wheel characteristic feel. 
Despite the proven and significant increase of MS from ND to RD (Figure \ref{fig:RC_MS_box}), this increase seems to not be relevant with the remote drivers' subjective assessment based on the above remarks.
This is another evidence of the greater complexity in remote drivers' steering feel and behaviour, which needs to be further investigated. 

Based on Table \ref{tab:RSquareOfCorrelation}, the overall driver feel assessment (SA01) illustrates decent to low correlation with RC for both ND and RD scenarios (R$^2$ = $\sim$ 0.20 and 0.16, respectively).
Despite these not satisfactory correlation levels, the fitted curves increasing - decreasing patterns (Figure \ref{fig:RC_MS_SA01_S}) are consistent with the UP and LB curves (fitted curves with 95$\%$ confidence levels).
However, the pattern of the fitted curves contradicts between ND and RD. 
In ND, the improvement of the overall driver feel assessment (SA01), leads to the deterioration of occupants' ride comfort. 
This conflicting relation is widely discussed in the literature \cite{papaioannou2022multi}, but not often captured.
In conventional vehicles, the driver receives cues/feedback from the vehicle that it responds intuitively to the driver’s commands (steering, throttle and brake inputs).
The optimization of these cues is mostly contradicting to motion comfort enhancement.
On contrary to this expected behaviour, in RD, RC and SA01 have a straight forward relation, where the improvement of drivers' overall feel leads to the comfort improvement. 
Thus, the development of remote control systems will potentially be less complicated with regards to comfort compared to conventional vehicles. 
The low correlation levels of SA01 with MS both in ND and RD, does not allow the identification of the same conflicting relation. 

The above mentioned contradictory behaviour is also identified in Level 2 SA metrics. 
More specifically, the improvement of the safety assessment (SA10) and the steering wheel characteristic feel (SA20, Figure \ref{fig:RC_MS_SA20_S}) leads to the deterioration and the improvement of ride comfort in ND and RD, respectively.
As proven before, in ND, the more perceived safety (i.e., increasing SA10) or realistic steering wheel characteristic feel (i.e., increasing SA20), the more confidence will be provided to the drivers (SA30).
This confidence increase is translated into a more aggressive driving including more steering velocity and throttle variations \cite{Lin2022a}.
Thus, more discomfort and MS symptoms are provoked to occupants (Figure \ref{fig:RC_MS_box}).
However, in RD, the improvement of confidence and control, which might occur from the improvement of SA10 and SA20, is translated to more situational awareness.
The remote drivers enhanced perceived situational awareness due to more confidence leads to a more comfortable and smooth ride.


Regarding the Level 3 SA metrics, the safety assessment related metrics are the ones to correlate more with RC both in ND and RD. 
More specifically, both the steering feedback support to control the vehicle (SA11) and the communication of the vehicle behaviour (SA12) displayed similar levels of correlation (R$^2$ = 0.24-0.27) with RC in ND and RD.
According to Figure \ref{fig:RC_MS_SA12_S}, the improvement in the steering feedback support or communication provokes more discomfort in ND, whereas it enhances comfort in RD.
As mentioned before, the increase of the feedback support or the communication in ND might provide more confidence to the driver. 
The feedback provides the drivers information about how the vehicle intuitively reacts to their inputs and allows them to adapt their steering, which might lead eventually to a more aggressive driving style.  
On the other hand, the remote driver adopts a more smooth and safe driving style with the the higher feedback support, which is in alignment with previous conclusions.

As far as motion sickness is concerned, in ND, it illustrates significant correlations with the Level 3 metrics, even reaching R$^2$ values larger than 0.60 (SA31 and SA32). 
On contrary, in RD, the values are decreased either to no correlation or by 50$\%$ (SA31 and SA32).
This does not allow any concrete conclusions to be extracted for most of the metrics, but interesting remarks can be highlighted for SA32 correlation with MS. 
According to Figure \ref{fig:ND_SA_ND_MS_SA31_S}-\ref{fig:RD_SA_RD_MS_SA31_S}, the contradictory relation between ND and RD is not anymore present when it comes to RC.
In both ND and RD, when the driver assesses the task completion as successful, the driving style has provoked more motion sickness both in RD and ND. 
This is because the drivers' confidence and control was high, resulting in a less smooth and more sickening drive. 





\subsection{Limitations}
This work presents the first in depth explorative analysis about the relation of motion comfort and driver steering feel in teleoperation by exploiting data collected from a partially published novel experiment \cite{Lin2022a}.
Despite the significant remarks extracted, there are certain limitations that the authors would like to highlight. 
Firstly and most importantly, the conclusions extracted are focused on the current experiment and its design. 
This relation could be modified if richer visual or audio cues are provided to the remote drivers improving their situational awareness. 
Secondly, the data used for this analysis are limited due to the complexity of the experiment. 
Only five drivers were employed to assess the steering feel both in normal and remote driving.
At each scenario, the drivers tested and assessed three different steering controllers, which resulted in 15 cases per scenario. 
To overcome the limited number of drivers, the authors considered the different controllers as different cases. 
For the authors to proceed in such assumption, they tested the consistency of the driving behaviour at each case (Figures \ref{fig:RC_MS_box}-\ref{fig:DT_SV_DS_box}).
Thirdly, the last limitation is that the experiment had limited the RCV-E maximum velocity around 18 km/h. 
The drivers, either in the normal or remote driving scenario, reached it relatively easy not allowing significant throttle variations. 
However, the results are of high value for slalom conditions and pave the path for further experiments that will allow the consideration of more longitudinal dynamics.


\section{Conclusions}

To sum up, this work explored the relation of motion comfort with remote drivers' feel and paves the path for properly designing steering feedback in remote driving with consideration of motion comfort.
Human drivers tested the steering feel both during normal and remote driving with three different steering feedback models. 
Drivers' feel subjective assessment is conducted through questionnaires, while objective metrics are also calculated to acquire comprehensive data regarding drivers' behavior or performance. 
Occupants' motion sickness (MS) and ride comfort (RC) are assessed with objective metrics. 
Based on the discussion, the following conclusions were extracted: 

\begin{itemize}
    \item The subjective driver feel assessment both in normal and remote driving could be simplified, since existing metrics in the questionnaire provide similar information. 
    For example, the overall driver feel assessment in normal driving could be evaluated based on the Level 2 questions related with safety, steering wheel characteristic feel and the confidence $\&$ control, whereas in remote driving only safety and confidence $\&$ control assessment can be used. 
    At the same time, the importance of the steering wheel characteristic feel is deteriorated in remote driving compared to normal driving. 
    Furthermore, the remote drivers' confidence and control is mainly affected by their perception of safety (steering feedback support and communication of vehicle behaviour), whereas in normal driving the steering wheel characteristic feel (level of feedback force and steering wheel returnability) and the confidence $\&$ control are also affecting it.     
    \item Motion sickness increases in average around 26 $\%$ from normal to remote driving, while steering velocity also increases around 25 $\%$. 
    These increases were consistent regardless the driver or the steering feedback controller. 
    At the same time, ride comfort and throttle input do not increase significantly from normal driving to remote driving.
    \item The correlation of motion sickness and ride comfort with the steering velocity is increased from normal to remote driving, whereas their correlation with the throttle variations is decreased. 
    Overall, the steering and throttle behaviour jointly affect the incidence of motion sickness and ride discomfort in normal driving, while in remote driving the steering behaviour is the dominant factor for the occurrence of both.    
    \item Subjective driver feel is more correlated with motion sickness than ride comfort in normal driving as expected due to the known impact of driving behaviour on motion sickness. 
    Meanwhile, there is no correlation in remote driving. 
    This is another evidence of the greater complexity in remote drivers' steering feel and behaviour, which needs to be further investigated. 
    Furthermore, in remote driving, ride comfort and driver feel have a straightforward relation, where the enhancement of drivers' overall feel leads to the comfort enhancement. 
    Hence, the development of remote control systems could be less complicated than conventional vehicle systems, where there is conflict between motion comfort and driver feel as outlined also in this work. 
\end{itemize}

Further work is in progress to subjectively assess motion comfort in remotely driven vehicles, validating the outcomes of this paper. 



\section*{Acknowledgment}

The authors would like to thank the following funders for supporting the work. At KTH Royal Institute of Technology, the REDO FFI project (grant number 2019-03068) and the Centre for ECO2 Vehicle Design (grant number 2016-05195), which both are funded by Vinnova (the Swedish innovation agency). Furthermore, the work has also been supported by  TRENoP (Transport Research Environment with Novel Perspectives) at KTH Royal Institute of Technology. 

\ifCLASSOPTIONcaptionsoff
  \newpage
\fi

\bibliography{Driver_Feel}
\bibliographystyle{IEEEtran}

%

\begin{IEEEbiography}[{\includegraphics[width=1.1in,height=1.25in,clip,keepaspectratio]{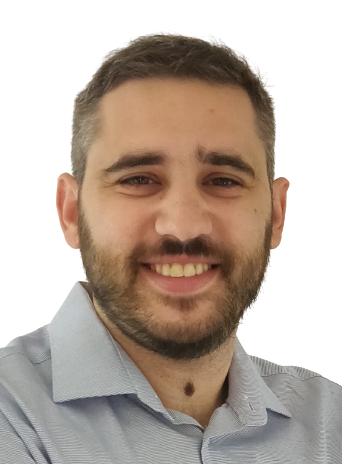}}]%
{Georgios Papaioannou} received the Ph.D. degree from the National Technical University of Athens (NTUA), Greece, in 2019, which received an award regarding its innovation and impact. He is currently an Assistant Professor on motion comfort in AVs at TU Delft, after conducting postdoctoral research at KTH Royal Institute of Technology in Sweden and Cranfield University in U.K. His research interests include motion comfort, seat comfort, postural stability, human body modelling, automated vehicles, motion planning, optimisation and control.
\end{IEEEbiography}

\begin{IEEEbiography}[{\includegraphics[width=1.1in,height=1.25in,clip,keepaspectratio]{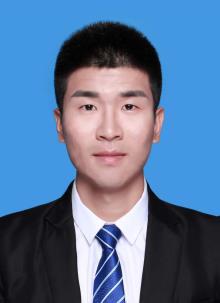}}]%
{Lin Zhao} was born in Hebei, China. He received the  M.S. degree in automotive engineering from Chongqing University, Chongqing, China, in 2019. He is currently
working toward the Ph.D. degree with the Department of Engineering Mechanics, School of Engineering Sciences, KTH Royal Institute of Technology, Stockholm, Sweden.
From 2017 to 2018, he was a visiting student in state key laboratory of automotive safety and energy, Tsinghua University, Beijing, China. From 2019 to 2020, he was a research engineer in Intelligent Vehicle R\&D Center of Gelly Automobile.
His research interests include teleoperation, autonomous vehicle, vehicle system dynamics and control, and dynamic game theory.
\end{IEEEbiography}

\begin{IEEEbiography}[{\includegraphics[width=1.1in,height=1.25in,clip,keepaspectratio]{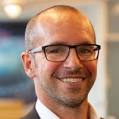}}]%
{Mikael Nybacka} received the Mechanical Engineer- ing degree in 2005 at Luleå University of Technol- ogy where he also received the Ph.D. degree. He currently works at KTH Royal Institute of Technol- ogy as an Associate Professor in Vehicle Dynamics.
His research focus is vehicle validation and driver vehicle interaction and also various aspects concerning over-actuated vehicles, e.g. design of over-actuation and urban vehicle concepts, control of autonomous vehicles and fault-tolerant control. 
\end{IEEEbiography}

\begin{IEEEbiography}[{\includegraphics[width=1.1in,height=1.25in,clip,keepaspectratio]{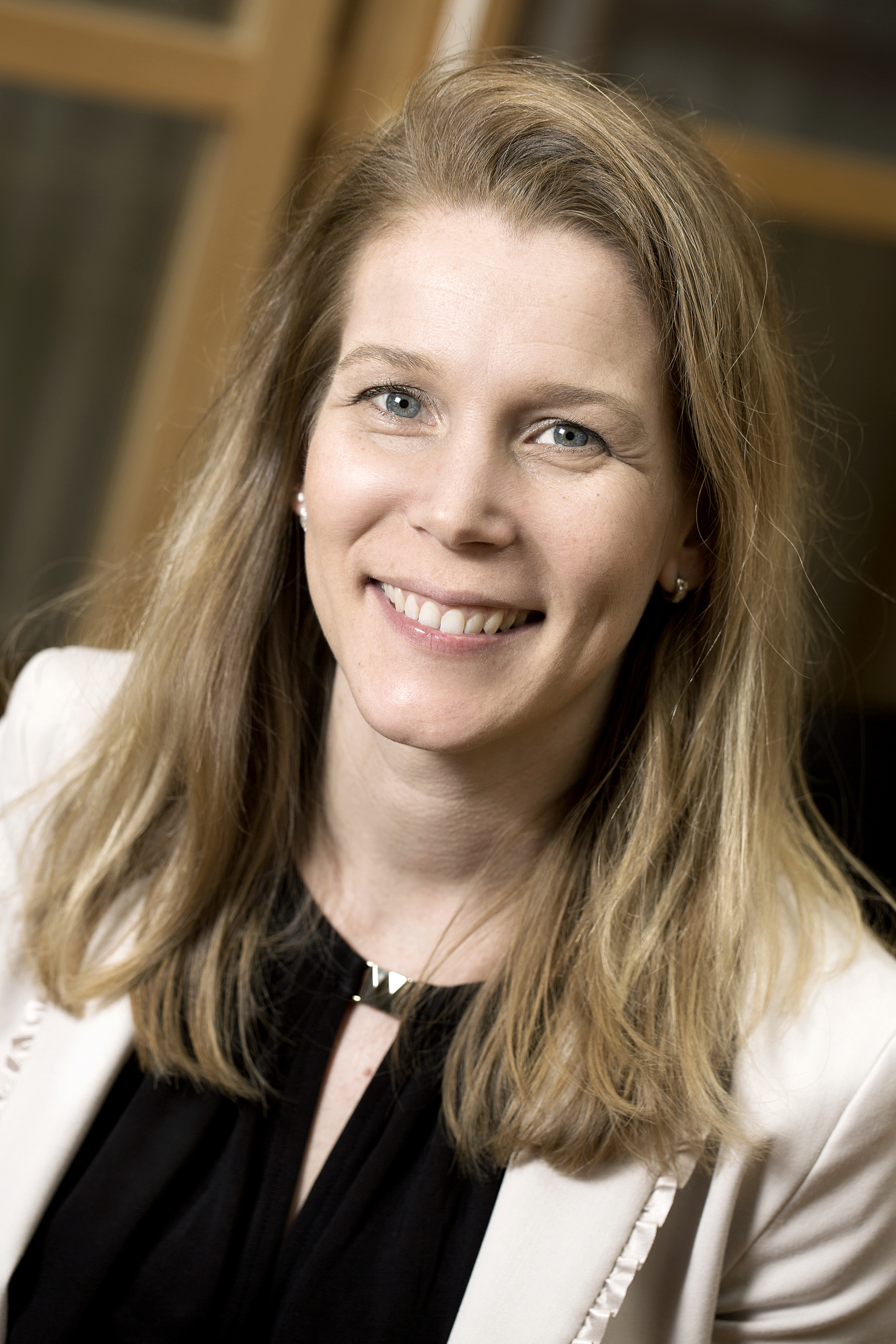}}]%
{Jenny Jerrelind} received her MSc degree in Engineering Physics from Luleå University of Technology, Sweden in 1998 and her PhD degree in Vehicle Dynamics from KTH Royal Institute of Technology, Sweden in 2004. She is working as an Associate Professor in Vehicle Dynamics at the Department of Engineering Mechanics, KTH Royal Institute of Technology, Sweden. Her research interest is vehicle dynamics with special focus on tyre modelling, mitigation of motion sickness, over-actuated vehicles and vehicle energy efficiency. 
\end{IEEEbiography}

\begin{IEEEbiography}[{\includegraphics[width=1.1in,height=1.25in,clip,keepaspectratio]{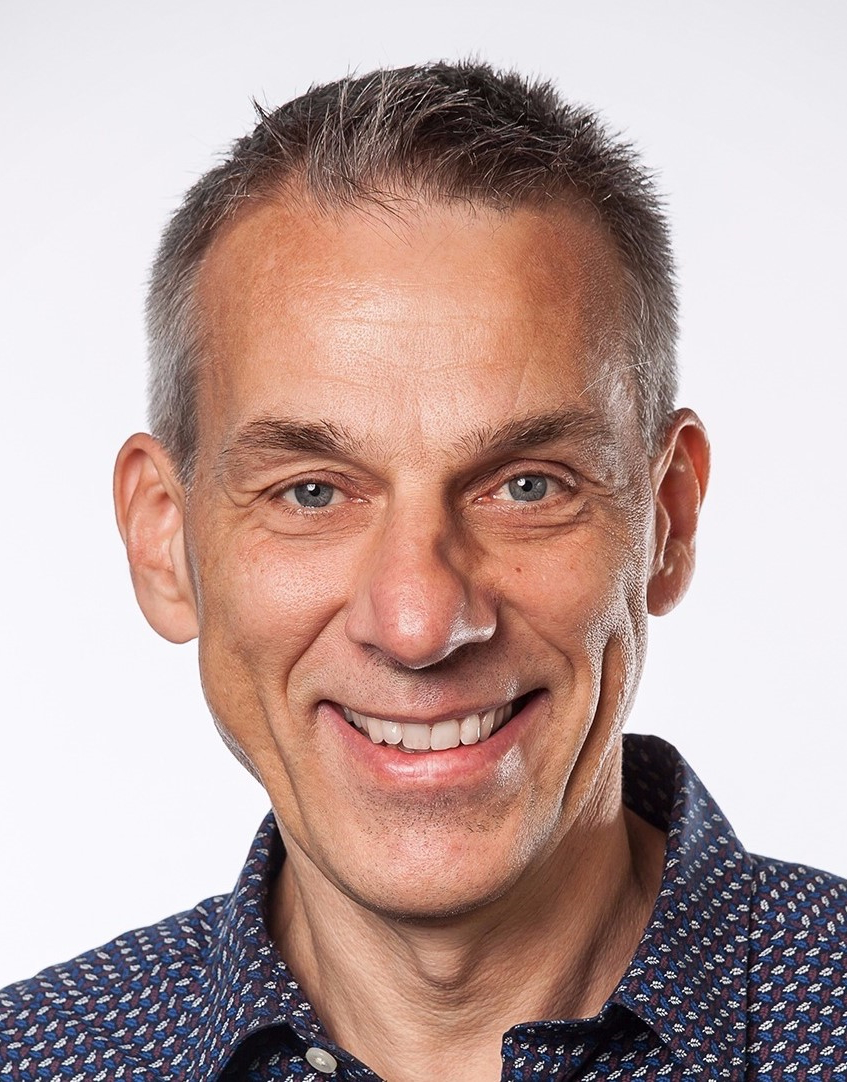}}]
{Riender Happee} received the Ph.D. degree from TU Delft, The Netherlands, in 1992.
He investigated road safety and introduced biomechanical human models for impact and comfort at TNO Automotive (1992-2007). Currently, he investigates the human interaction with automated vehicles focusing on safety, motion comfort and acceptance at the Delft University of Technology, the Netherlands, where he is full Professor.
\end{IEEEbiography}

\begin{IEEEbiography}[{\includegraphics[width=1.1in,height=1.25in,clip,keepaspectratio, angle=-90]{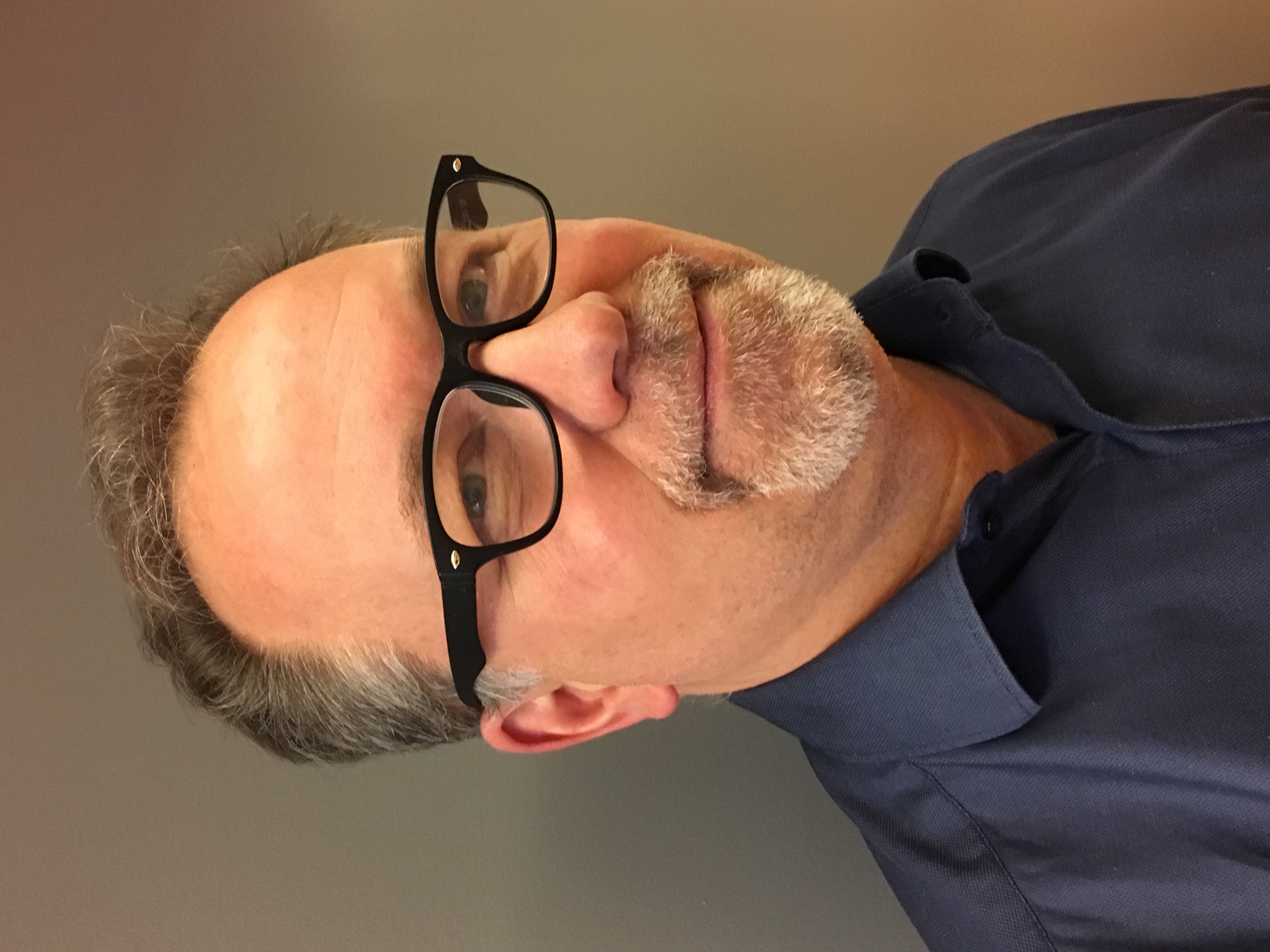}}]%
{Lars Drugge} received his M.Sc. degree in Mechanical Engineering in 1994 and his Ph.D. in Computer Aided Design in 2000 from Luleå University of Technology, Sweden. He is working as an Associate Professor in Vehicle Dynamics at the Department of Engineering Mechanics, School of Engineering Sciences, KTH Royal Institute of Technology, Stockholm, Sweden. Assoc. Prof. Drugge is currently head of the Vehicle Dynamics, Rail Vehicles and Conceptual Vehicle Design group. His areas of research include vehicle dynamics of over-actuated vehicles, tyre modelling, driver vehicle interaction and driving simulators.
\end{IEEEbiography}

\end{document}